\documentclass[runningheads]{llncs}
 
\usepackage{eccv}

\usepackage{eccvabbrv}

\usepackage{graphicx}
\usepackage{booktabs}

\usepackage[accsupp]{axessibility}  

\usepackage{amsmath}
\usepackage{booktabs}
\usepackage{graphicx}
\usepackage{multirow}
\usepackage{caption}
\usepackage{array}
\usepackage{makecell}
\usepackage{tabularx}
\usepackage{xcolor}
\usepackage[table]{xcolor}
\usepackage[dvipsnames]{xcolor}

\usepackage[percent]{overpic}
\usepackage{pict2e}
\usepackage{wrapfig}
\usepackage{algorithm}
\usepackage{algorithmic}
\usepackage{tocloft}
\usepackage{etoc}

\makeatletter
\def\subparagraph{\@startsection{subparagraph}{5}{\parindent}%
  {-12\p@ \@plus -4\p@ \@minus -4\p@}{-0.5em}%
  {\normalfont\normalsize\bfseries}}
\makeatother

\usepackage{titlesec}
\usepackage{titletoc}

\newcommand{\sys}{\ensuremath{\mathrm{APT}}\xspace}
\newcommand{\syss}{\ensuremath{\mathrm{APT^*}}\xspace}

\definecolor{ours}{HTML}{F4F8FF}

\definecolor{bestcolor}{HTML}{CB0000} 
\definecolor{sbestcolor}{HTML}{0054A6} 

\newcommand{\best}[1]{\textcolor{bestcolor}{\textbf{#1}}}
\newcommand{\sbest}[1]{\textcolor{sbestcolor}{\underline{#1}}}

\usepackage{hyperref}

\usepackage{orcidlink}

\begin{document}

\title{\textsc{APT}: Anchor-aligned Perturbations for Tamper Localization in Fully Regenerated Images}

\titlerunning{Anchor-aligned Perturbations for Tamper Localization in FR Images}

\author{Suhyeon Ha\and
Woo Jae Kim\and
Joonsung Jeon\and
Sooel Son\textsuperscript{*}\and
Sung-eui Yoon\textsuperscript{*}}

\authorrunning{S.~Ha et al.}

\institute{KAIST \\
\email{\{suhyeon.ha, wkim97, mikeraph, sl.son, sungeui\}@kaist.ac.kr}
\url{https://suhyeonha.github.io/APT}}

\maketitle

\begingroup
\renewcommand{\thefootnote}{*}
\footnotetext{Co-corresponding authors}
\endgroup

\begin{abstract}
  \vspace{-7pt}
  Proactive tamper localization embeds an imperceptible signal into an image prior to distribution, enabling pixel-level manipulation detection. Existing methods assume a spliced (SP) setting, where synthesized regions are composited onto the original background, leaving embedded signals intact. However, real-world diffusion-based inpainting operates in a fully regenerated (FR) setting, where the entire image undergoes denoising, disrupting background signals and rendering existing frameworks ineffective. We propose \sys, a semi-fragile latent-space perturbation that embeds a dense, vector-wise localization signal. By aligning each spatial feature vector toward a fixed anchor direction, \sys localizes tampering via the alignment disparity between synthesized foreground and anchor-aligned background  features after inpainting. The proposed hard negative mining loss and noisy perturbation branch further enforce uniform alignment. Experiments on COCO demonstrate that \sys  achieves an FR IoU of 0.92, outperforming the strongest baseline (WAM, 0.84), while existing methods collapse to near-random performance (AUC $\approx$ 0.5), establishing \sys  as a practical forensic framework generalizable across tampering types unknown at test time.
  \vspace{-7pt}
\end{abstract}

\begin{figure}[h]
  \centering
  \includegraphics[width=0.99\linewidth, trim={0.1cm 5.1cm 2.3cm 0cm}, clip]{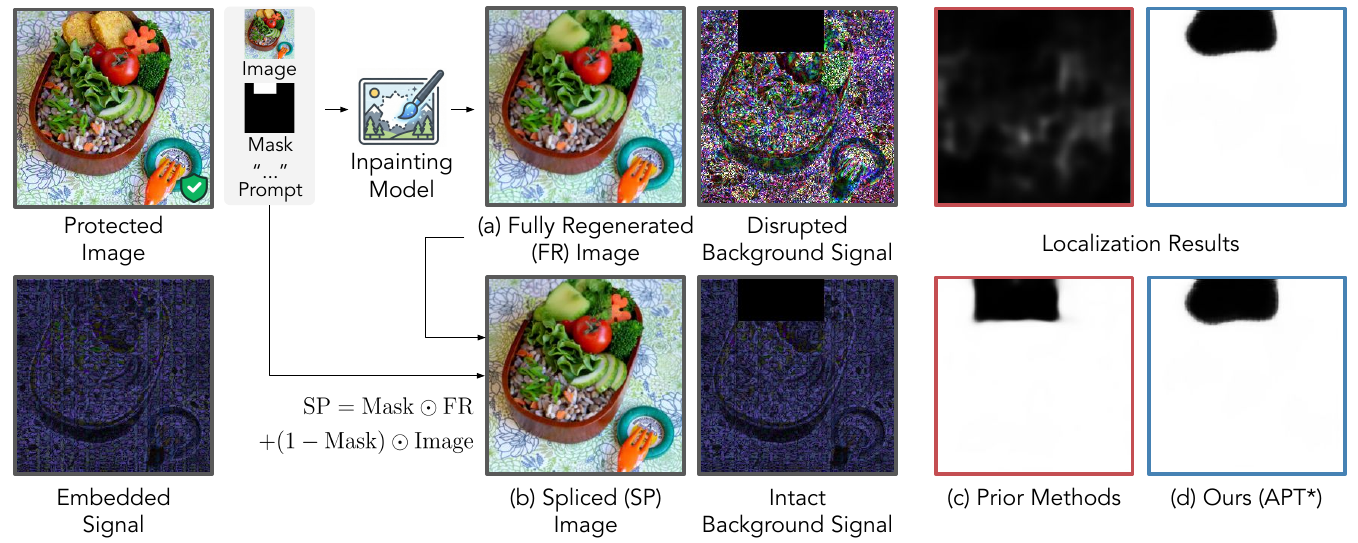}
  \caption{Vulnerability of proactive localization under fully regenerated (FR) inpainting. In the spliced (SP) setting (b), assumed by existing methods, synthesized regions are composited onto the original background, leaving the embedded signal intact. In contrast, the FR setting (a) uses the raw inpainting output directly, disrupting the background signal across the entire image. While prior methods (c) struggle to localize tampered regions under FR due to severe background signal disruption, our method (d) produces accurate localization masks under both FR and SP settings. Background signals are visualized as $10\times(|tamper-original|\odot(1-mask))$.}
  \vspace{-7pt}
  \label{fig:fig1}
\end{figure}

\section{Introduction}
\label{sec:intro}

Recent advances in diffusion-based inpainting~\cite{ldm, brushnet, controlnet, hdpainter} enable highly photorealistic, text-guided image manipulation.
These generative capabilities substantially lower the technical barrier for fraudsters, facilitating the large-scale production and dissemination of deceptive visual forgeries on social media, particularly during high-stakes global crises~\cite{AP2023, Guardian2023}. Such misuse transforms inpainting from a creative tool into an instrument for information manipulation.
To counter this threat, proactive tamper localization has emerged as a defensive scheme. The core idea is to embed an imperceptible signal into an image prior to distribution. Upon suspected manipulation, verifiers can analyze the protected image and localize the altered regions.

Existing proactive localization methods~\cite{editguard, wam, omniguard, stableguard} rely on an optimistic adversary model that assumes splicing (SP): after synthesizing a forged region, the adversary copies the forged patch back onto the original image as a post-processing step (\cref{fig:fig1}). Under this assumption, the background remains unchanged and the embedded signal outside the edited region is preserved, enabling reliable localization.
However, this assumption is broken by diffusion-based editing in its native form.
The raw output of diffusion models is a fully regenerated (FR) image; the entire image undergoes the noise sampling and denoising process to blend the synthesized foreground with the background.
Accordingly, the embedded signal in the background is also impaired, even when the semantic content outside the edited region appears intact.
While existing studies~\cite{sagi, tgif} note that FR images are generally harder to detect than SP images, no prior work in proactive localization has addressed this practical and challenging FR setting.

Addressing this gap requires rethinking what properties a localization signal must satisfy to be effective in the FR setting.
Diffusion-based inpainting models synthesize the target foreground based on a text prompt while reconstructing the original background. Successful localization under such generative processes thus requires a dual property: the signal must remain robust to background regeneration yet fragile to foreground synthesis.

However, existing proactive methods~\cite{wam, omniguard, stableguard} struggle to satisfy this requirement, particularly the robustness to background regeneration, due to two primary limitations. 
First, their localization embeddings are fragile to both the VAE encoding-decoding process and pixel regeneration used in diffusion pipelines. As a result, signals embedded in background regions become unstable, rendering them unreliable as localization cues in the FR setting.
Second, they often rely on global embeddings that lack explicit spatial correspondence between the signal and its location, making it difficult to accurately map pixel-level tampering.

To address these challenges, we propose \sys, a semi-fragile perturbation framework that embeds a robust vector-wise localization signal in the VAE latent space.
\sys optimizes a latent perturbation that drives each spatial latent vector toward a predefined anchor vector, producing vector-wise alignment at every spatial location rather than relying on a single global embedding.
Since we sample this anchor vector from the unit hypersphere independently of the generative model's learned distribution, feature vectors synthesized by the diffusion model naturally deviate from this anchor direction.
In contrast, during diffusion-based inpainting, background regions are reconstructed by heavily conditioning on the original latent representation, which already carries the anchor-induced alignment.
This foreground-background alignment disparity yields a reliable localization signal that differentiates tampered and protected regions after inpainting.
To further stabilize the signal, we introduce a hard negative mining loss that suppresses weakly aligned spatial features and a noisy perturbation branch that improves robustness of the embedded signal against VAE-induced distortions during background regeneration.

\sys achieves consistent localization performance across both spliced and fully regenerated tampered images. At verification, the vector-wise detection map is directly upsampled to produce a pixel-level mask (\sys), while an optional lightweight decoder can be employed to further refine spatial boundaries (\syss). 

We compare \sys with state-of-the-art tamper localization methods across both SP and FR settings. While existing methods perform well in SP settings, they suffer severe performance drops in IoU ranging from 0.14 to 0.93 when faced with FR manipulations. Specifically, StableGuard~\cite{stableguard} and OmniGuard~\cite{omniguard} collapse to near-random AUC levels ($\approx$ 0.5), whereas \sys  maintains a robust AUC of 0.97. Furthermore, \syss surpasses the top-performing baseline (WAM~\cite{wam}, 0.84), with an IoU of 0.92, effectively bridging the gap in FR tamper localization.

Our contributions are as follows: 
\vspace{-2pt}
\begin{itemize}
    \item We present the first systematic evaluation of proactive tamper localization under fully regenerated (FR) inpainting. We reveal a critical vulnerability in existing methods, suffering severe performance degradation on FR forgeries.
    \item We propose \sys, a novel semi-fragile perturbation framework that optimizes a dense, vector-wise anchor signal in the VAE latent space, where the foreground-background alignment disparity serves as the localization cue. A hard negative mining loss and a noisy perturbation branch further enforce spatially uniform alignment, enabling robust localization across both SP and FR manipulations.
    \item Experiments on COCO demonstrate that \sys significantly outperforms state-of-the-art baselines under the challenging FR manipulations while maintaining competitive performance on SP images. This establishes \sys as a practical tamper localization framework for real-world scenarios where the manipulation type is unknown at test time.
\end{itemize}

\section{Related Work}

\subsection{Diffusion-based Inpainting}
\label{sec:related_inpainting}
Diffusion-based inpainting~\cite{ldm, repaint, hdpainter, controlnet, brushnet} takes as input an image, a binary mask, and a text prompt, and synthesizes content within the masked region so that the generated foreground is semantically consistent with the prompt and visually coherent with the surrounding image. 
To preserve background consistency, recent inpainting models integrate spatial context into the denoising process through several mechanisms.
For example, \cite{bld, repaint, hdpainter} merge the denoised foreground with the noised original background according to the provided mask.
\cite{ldm, hdpainter} concatenate the mask, the masked background image, and the noisy latent along the channel dimension and feed them as inputs to the U-Net~\cite{unet} in the inpainting models.
Also, \cite{controlnet, brushnet} inject mask-conditioned features from a secondary branch into a frozen U-Net feature through zero convolutions.  

With these background-preserving mechanisms, modern inpainting models generate visually coherent outputs without requiring any post-processing.
Consequently, adversaries are able to distribute the raw outputs of these models, in which the entire images are regenerated through the diffusion process.
Therefore, tamper localization remains a critical challenge especially when synthesized images do not involve explicit splicing--a setting we refer to as fully regenerated (FR).

\subsection{Tamper Localization}
Traditional passive localization methods ~\cite{mvssnet, hifinet,trufor,sida,clip_ifdl,diffforensics} identify tampered regions by analyzing multi-view boundary artifacts~\cite{mvssnet}, extracting hierarchical fine-grained features~\cite{hifinet}, detecting noise anomalies~\cite{trufor}, or leveraging large multimodal models~\cite{sida}. While effective against manual editing techniques (\eg copy-move, splicing), these passive methods struggle to detect photorealistic manipulations produced by advanced generative models~\cite{editguard, omniguard}.

To address the limitations of passive methods, recent methods proactively embed localization signals before alterations.
MaLP~\cite{malp} learns proactive spatial templates by incorporating a generative model directly into the framework. EditGuard~\cite{editguard} utilizes invertible networks in wavelet domain to hide a secret image, while WAM~\cite{wam} treats watermarking as a localized segmentation task. OmniGuard~\cite{omniguard} extends EditGuard by injecting VAE compression and degradation-aware extractor. Furthermore, StableGuard~\cite{stableguard} integrates watermarking during latent diffusion generation, utilizing a mixture-of-experts network for forensic extraction.

Despite recent advances in proactive tamper localization, existing methods struggle to generalize to FR inpainting due to two primary limitations. First, methods such as WAM~\cite{wam} and OmniGuard~\cite{omniguard} embed signals in the pixel or frequency domain, making them inherently susceptible to the VAE bottleneck and background regeneration. Second, methods such as StableGuard~\cite{stableguard} apply a single global signal across the entire image, lacking a direct spatial correspondence between the signal and its location. Without locally dense signals, mask predictors tend to overfit to visual cues such as splicing boundaries and collapse entirely from the global distortion under FR settings. This gap motivates a framework that combines latent-space perturbation optimization with spatially dense signal embedding.

\section{Preliminary}

\subsection{Spliced vs. Fully Regenerated Images}
The key difference between spliced (SP) and fully regenerated (FR) manipulations lies in whether the unmasked background is preserved.
Let $x' = \mathcal{G}(x, m, c)$ denote the output of the generative model $\mathcal{G}(\cdot)$ conditioned on image $x$, mask $m$ ($m=1$ for the editing region), and prompt $c$.
For an SP image, $x'_\text{SP}$ is composited onto the original background using the binary mask $m$:
\begin{equation}
x'_\text{SP} = x' \odot m + x \odot (1 - m),
\end{equation}
where $\odot$ denotes the element-wise product.
In this case, the original background is strictly preserved, ensuring $x'_\text{SP} \odot (1 - m) = x \odot (1 - m)$.
Consequently, any proactive signals embedded in the background remain undisturbed.
In FR, the raw output from the generative model is used directly:
\begin{equation}
x'_\text{FR} = x'.
\end{equation}
Here, the entire image undergoes stochastic noise injection and iterative denoising. Even regions outside the masked area are re-synthesized. Thus, pixel-level equivalence no longer holds (\ie $x_\text{FR} \odot (1 - m) \neq x \odot (1 - m)$).

This discrepancy arises from multiple factors: (1) statistical disharmony between the replaced background and the model's own background generation, (2) generative priors that regularize pixel values toward the learned data distribution~\cite{repaint}, and (3) latent reconstruction errors in VAE-based inpainting models, which prevent exact recovery of original pixel-level details~\cite{ldm, brushnet}.
Therefore, existing proactive localization methods become vulnerable to adversaries who employ diffusion-native FR, necessitating a detection framework that remains robust under full-image regeneration.

\subsection{Anchor-aligned Proactive Perturbation}
To establish a proactive defense, we build upon the zero-bit watermarking framework introduced in \cite{ssl}, which determines the presence of a hidden mark.
Given an original image $\mathbf{x} \in \mathbb{R}^{H \times W \times 3}$, 
an imperceptible perturbation is added to generate a marked image $\tilde{\mathbf{x}} = \mathbf{x} + \boldsymbol{\delta}$. 
The perturbation is optimized via gradient descent to push an image feature vector $\mathbf{v}$ toward a predefined anchor vector $\mathbf{a}$. This process effectively constrains the image feature within a decision boundary known as a hypercone $\mathcal{C}$, which is formally defined as the set of vectors satisfying:
\begin{equation}
\mathcal{C} = \left\{ \mathbf{v} \mid \frac{\mathbf{v} \cdot \mathbf{a}}{\|\mathbf{v}\| \|\mathbf{a}\|} \ge \cos(\theta) \right\},
\end{equation}
where the feature is geometrically aligned within an angular distance $\theta$ from the anchor $\mathbf{a}$. During detection, an image is identified as marked if its extracted feature vector resides within $\mathcal{C}$. 

\section{Method}

\begin{figure}[tb]
  \centering
  \includegraphics[width=0.98\linewidth, trim={0cm 2.1cm 0cm 0cm}, clip]{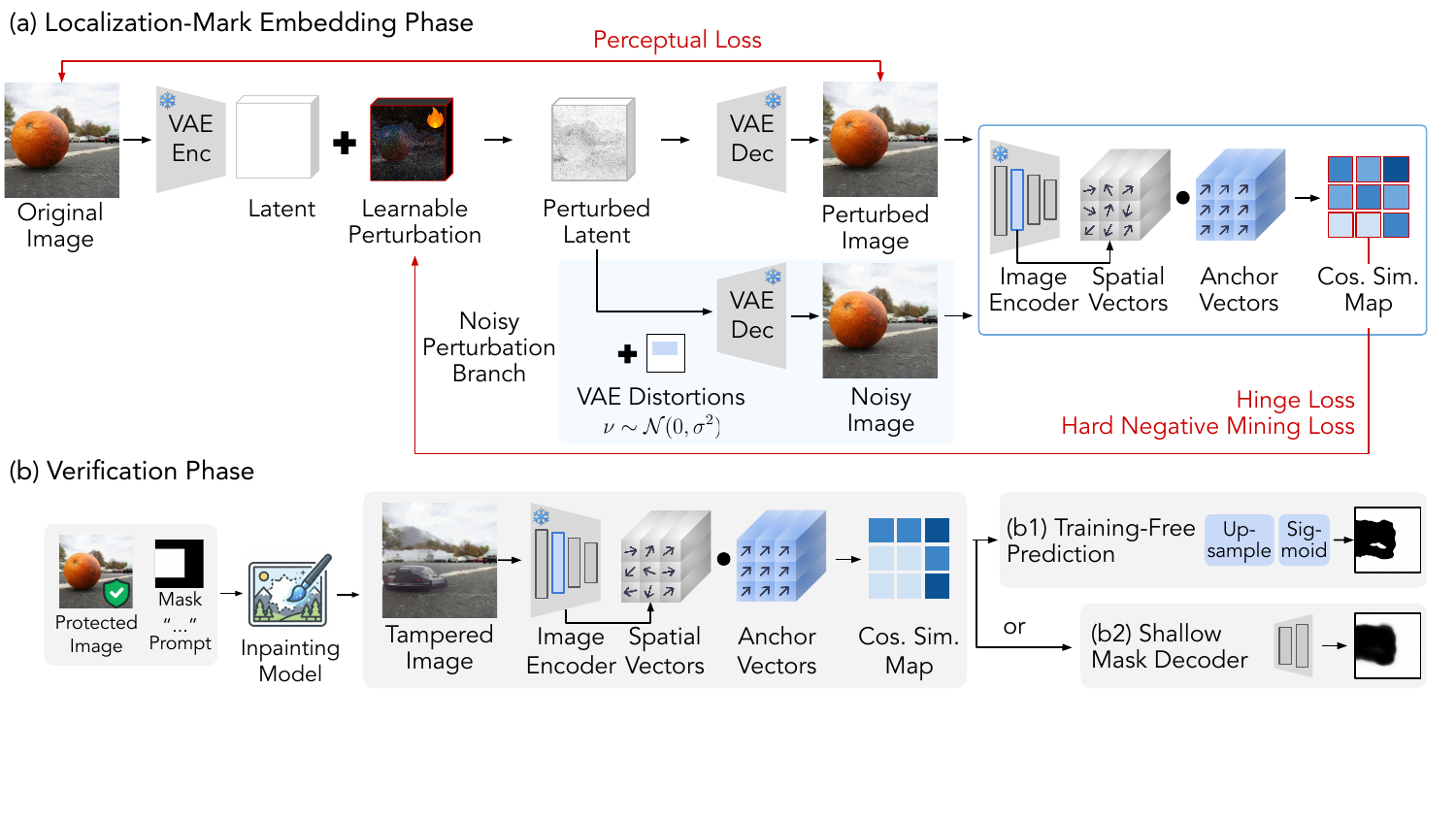}
  \caption{Overview of the \sys pipeline. (a) In the localization-mark embedding phase, a learnable perturbation is optimized in the VAE latent space to align spatial feature vectors toward predefined anchors. (b) In the verification phase, a per-vector cosine similarity map extracted from a potentially tampered image is passed to either (b1) a training-free predictor or (b2) a shallow mask decoder to produce a localization mask.}
  \label{fig:framework}
\end{figure}

We propose \sys, a semi-fragile proactive localization framework that embeds a dense, vector-wise signal via anchor alignment in latent space (\cref{method:perturb}).
\sys embeds an imperceptible perturbation into an input image such that, in the resulting protected image, background features remain consistently aligned with predefined anchors even after full regeneration (FR).

\noindent \textbf{Overview.}
\cref{fig:framework} illustrates the overall pipeline of \sys, which consists of two phases: localization-mark embedding and verification.
In the localization-mark embedding phase, \sys begins with an input image under protection and predefined anchor vectors.
The perturbed latent, along with noisy variants of the perturbation, is decoded through the VAE decoder to produce two perturbed image variants. These images are subsequently processed by an image encoder that outputs spatial feature vectors.
\sys optimizes the learnable perturbation to align these spatial features toward predefined anchor directions.
The key insight is that anchor-aligned background features persist--even under FR--whereas synthesized foreground regions follow the generative model's native distribution and exhibit low alignment with the anchor.
This distributional separation enables precise, pixel-level localization without assuming strict background preservation.
For this, we propose (1) a hard negative mining loss and (2) a noisy perturbation branch that enable spatially uniform alignment (\cref{method:robustness}).
We also propose training-free and shallow mask detectors tailored to SP and FR forgeries (\cref{method:localization}) that  do not require any tampered or optimized samples.

The verification phase takes a potentially tampered image as input, extracts per-vector anchor alignment maps, and produces a pixel-level localization mask via either training-free prediction or a shallow mask decoder.

\subsection{Anchor-aligned Perturbation for Localization}
\label{method:perturb}

\noindent \textbf{Embedding vector-wise perturbation.}
Given an original image $\mathbf{x} \in \mathbb{R}^{H \times W \times 3}$, \sys directly optimizes the perturbation in the VAE latent space to enhance the robustness against VAE compression and pixel regeneration. Given an initial latent $\mathbf{z} = \mathcal{E}_{\text{vae}}(\mathbf{x})$ extracted 
from the original image using a pretrained VAE encoder $\mathcal{E}_{\text{vae}}$, we add a learnable latent perturbation $\boldsymbol{\delta}$ to construct the perturbed latent $\tilde{\mathbf{z}} = \mathbf{z} + \boldsymbol{\delta}$.
The perturbed latent $\tilde{\mathbf{z}}$ is decoded via $\mathcal{D}_{\text{vae}}$ to obtain a temporary perturbed image $\tilde{\mathbf{x}}_{\text{perturb}} = \mathcal{D}_{\text{vae}}(\tilde{\mathbf{z}})$, which is passed through $\mathcal{E}_{\text{img}}$ to extract a dense feature map $\mathbf{F} \in \mathbb{R}^{H' \times W' \times d}$. We denote each feature vector at location $(i, j)$ as a \textit{spatial vector} $\mathbf{f}_{ij} \in \mathbb{R}^{d}$, which corresponds to a $s \times s$ pixel region in the input image with $s = H/H' = W/W'$.

To serve as a predefined and fixed reference for tamper localization, each anchor vector is sampled from a Rademacher distribution and normalized to have unit norm, constructing $\mathbf{a}_{ij} \in \left\{-\frac{1}{\sqrt{d}}, 
+\frac{1}{\sqrt{d}}\right\}^d$, approximating a uniform distribution on the unit hypersphere.
Under this construction, the cosine similarity between any clean feature and $\mathbf{a}_{ij}$ is statistically bounded near zero~\cite{vukotic, ssl}, indicating the absence of any intentional alignment. The degree of alignment between $\mathbf{f}_{ij}$ and its corresponding anchor $\mathbf{a}_{ij}$ is then measured by their cosine similarity $s_{ij} = \frac{\mathbf{f}_{ij}^\top \mathbf{a}_{ij}}{\|\mathbf{f}_{ij}\| \|\mathbf{a}_{ij}\|}$.

The perturbation $\boldsymbol{\delta}$ is then optimized to embed the vector-wise localization signal by enforcing $s_{ij}\geq\tau$ across all spatial locations via a hinge loss $\mathcal{L}_{\text{hinge}} = \frac{1}{H'W'} \sum_{i,j} \max(0, \tau - s_{ij})$, which drives each spatial vector to align within the target hypercone margin $\tau \in (0, 1)$. Perceptual losses (\ie $\mathcal{L}_{\text{PSNR}}$ and $\mathcal{L}_{\text{LPIPS}}$) are also applied to minimize perceptual distortion between the clean and perturbed images.

\noindent \textbf{Our objective.}
Our anchor-aligned perturbation $\boldsymbol{\delta}$ steers image features toward a predefined anchor direction, inducing intentional alignment that serves as a localization signal.
This perturbation is designed to be impaired in synthesized foreground regions while remaining stable in background regions.
Diffusion models generate foreground content by sampling from a learned data distribution derived from natural images~\cite{repaint}. Consequently, synthesized foreground regions tend to follow this clean distribution and exhibit low cosine similarity with the anchor. This occurs because the anchor is sampled as a unit vector from the hypersphere, making it statistically unlikely to align with spatial feature vectors extracted from generated images.
In contrast, inpainting models reconstruct the background by injecting the original latent representation.
This objective preserves the anchor-induced alignment in background regions, even under regeneration.
The resulting disparity--aligned background versus misaligned synthesized foreground--creates a measurable contrast in feature space that enables precise spatial localization. We empirically validate these properties in \cref{exp:fragility}.

\subsection{Hard Negative Mining Loss and Noisy Perturbation Branch}
\label{method:robustness}

\noindent
\textbf{Hard negative mining loss.} 
In \sys, the cosine similarity map is directly used as a localization signal, making spatially uniform alignment critical for reliable detection.
However, in general, features from the pretrained image encoder are non-uniformly distributed on the hypersphere~\cite{ssl}, causing the initial cosine similarity to vary across spatial locations.
We thus propose a hard negative mining loss to further penalize under-optimized spatial vectors.
Specifically, we exploit the loss on the top-$K$ spatial vectors with the highest per-vector hinge loss $l_{ij} = \max(0, \tau - s_{ij})$:
\begin{equation}
    \mathcal{L}_{\text{hard}} = \frac{1}{|\Omega_K|} 
    \sum_{(i,j) \in \Omega_K} l_{ij}
\end{equation}
where $\Omega_K$ denotes the index set of the top-$K$ spatial vectors sorted by $l_{ij}$ in descending order, with $K = \lceil \rho \cdot H'W' \rceil$ and $\rho$ being the sampling ratio.

\noindent
\textbf{Noisy perturbation branch.}
To enhance robustness of the embedded signal against VAE distortions under background regeneration, we introduce a noisy perturbation branch during optimization. A random binary rectangular mask $m_\text{rand}$ ($m_\text{rand}=1$ for noise-injected regions) is applied to inject Gaussian noise $\nu \sim \mathcal{N}(0, \sigma^2)$ into the perturbed latent $\tilde{\mathbf{z}}$, yielding a noisy latent:
\begin{equation}
\tilde{\mathbf{z}}_{\text{noisy}} = \tilde{\mathbf{z}} 
\odot (1 - m_\text{rand}) + (\tilde{\mathbf{z}} + 
\nu) \odot m_\text{rand}
\end{equation}
The noisy latent is decoded as $\tilde{\mathbf{x}}_{\text{noisy}} 
= \mathcal{D}_{\text{vae}}(\tilde{\mathbf{z}}_{\text{noisy}})$, 
and both $\mathcal{L}_{\text{hinge}}$ and 
$\mathcal{L}_{\text{hard}}$ are applied to the resulting 
noisy image. This strategy effectively improves robustness against latent-space distortions introduced by VAE regeneration.

\noindent
\textbf{Final objective function.} 
With the model parameters all frozen, the final objective function for updating the perturbation $\boldsymbol{\delta}$ is defined as:
\begin{equation}
\label{eq:full}
    \mathcal{L}_{\text{total}} = \sum_{\hat{\mathbf{x}} \in \{\tilde{\mathbf{x}}_{\text{perturb}},\, \tilde{\mathbf{x}}_{\text{noisy}}\}} \left[ \mathcal{L}_{\text{hinge}}(\hat{\mathbf{x}}) + \mathcal{L}_{\text{hard}}(\hat{\mathbf{x}}) \right] + \lambda_p \mathcal{L}_{\text{PSNR}} + \lambda_l \mathcal{L}_{\text{LPIPS}},
\end{equation}

where $\mathcal{L}_{\text{hinge}}(\hat{\mathbf{x}})$ and $\mathcal{L}_{\text{hard}}(\hat{\mathbf{x}})$ are computed on the cosine similarity map $\hat{s}$ extracted from $\hat{\mathbf{F}}$ via $\mathcal{E}_{\text{img}}$. The full objective jointly enforces spatially uniform anchor alignment across both the perturbed and noisy images, while minimizing perceptual distortion.
After iterative gradient updates, we enforce an $\ell_\infty$-norm constraint by clipping the pixel-domain perturbation within a budget $\epsilon_\delta$. The final perturbed image $\tilde{\mathbf{x}}$ is obtained as: $\tilde{\mathbf{x}} = \text{Clip}(\mathbf{x} + \text{Clip}(\tilde{\mathbf{x}}_{\text{perturb}} - \mathbf{x}, 
-\epsilon_\delta, \epsilon_\delta),\ 0, 1)$.

\subsection{Tamper Localization}
\label{method:localization}
At verification, given a test image $\tilde{\mathbf{x}}'$, we first extract its feature map via $\mathcal{E}_{\text{img}}$ and compute the per-vector cosine similarity map $s' \in \mathbb{R}^{H' \times W'}$. Average pooling is then applied to $s'$ to smooth per-vector variance, and the resulting map is passed to the detection function $\mathcal{D}$, which produces a pixel-level prediction mask $m' \in \{0, 1\}^{H \times W}$ (\ie $\mathcal{D}(s')=m'$) via either training-free prediction or a shallow mask decoder, as described below.

\noindent
\textbf{Training-free prediction.}
The detection function $\mathcal{D}$ processes the similarity map to produce a pixel-level prediction mask via bilinear interpolation, temperature scaling, and sigmoid activation $\sigma(\cdot)$:
\begin{equation}
    m' = \sigma\left(T \cdot 
    \text{Upsample}(s')\right),
\end{equation}
where $T$ is the temperature parameter. The values 1 and 0 in the predicted mask $m'$ represent perturbed and tampered regions, respectively. A binary localization mask is obtained by thresholding at $0.5$.

\noindent
\textbf{Shallow mask decoder.}
To achieve fine-grained localization, we introduce a lightweight shallow mask decoder. Since vector-wise prediction operates independently at each spatial location, it may produce spatially inconsistent masks with isolated false predictions.
The decoder addresses this by learning a locality prior that masked regions tend to be spatially contiguous. This allows to leverage surrounding context to suppress isolated errors and fill incomplete predictions.

The decoder $\mathcal{D}_{\theta}$, with learnable parameters $\theta$, is trained to predict a mask $m' \in \mathbb{R}^{1 \times H \times W}$ from a cosine similarity map $\hat{s}'$, supervised against a ground-truth (GT) mask $m_\text{GT}$ (1: perturbed, 0: tampered) by a joint loss combining binary cross-entropy (BCE) and Dice loss.
Since constructing a large-scale dataset of paired perturbed and tampered images is practically infeasible, we instead generate synthetic similarity maps from clean images alone.
Specifically, we construct a synthetic similarity map $\hat{s}'$ to simulate the expected cosine similarity distribution after inpainting, by adding alignment gain to background regions and imposing alignment collapse on foreground regions:
\begin{equation}
    \hat{s}' = \underbrace{(s' + \Delta_{\text{bg}})}_{\text{alignment gain}} 
    - \underbrace{\Delta_{\text{fg}} \cdot (1 - m_\text{GT})}_{\text{alignment collapse}} 
    + \eta,
\end{equation}
where $\Delta_{\text{bg}} \sim \mathcal{U}(\alpha_1\tau, \alpha_2\tau)$ models the alignment gain in perturbed regions, and $\Delta_{\text{fg}} \sim 
\mathcal{U}(\beta_1\tau, \beta_2\tau)$ models the alignment collapse in tampered areas. Here, $\eta \sim \mathcal{N}(0, 0.03^2)$ is additive noise that is introduced to inject per-vector variance in the similarity map. This decoder comprises two upsampling stages ($\times 2$ and $\times 4$), each comprising an upsampling layer followed by two conv-LayerNorm-ReLU blocks.

\section{Experiments}
\subsection{Implementation Details}
We optimize an image at resolution $H\times W=256\times256$ with $H'\times W'=32\times32$. We use DINOv3-ConvNeXt-Small~\cite{dinov3} as the image encoder with feature layer 1 ($d=192$), and the VAE from \texttt{sd-legacy/stable-diffusion-inpainting}~\cite{ldm}. The perturbation is optimized solely on the VAE latent space, without the U-Net. We optimize $\boldsymbol{\delta}$ using Adam~\cite{adam} with lr$=1.0$ over 150 steps, with $\lambda_p=0.1$, $\lambda_l=0.05$, $\epsilon_\delta=16/255$, $\tau=0.1$, $T=5.0$ and $\rho=0.1$. The noise scale $\epsilon$ for the noisy branch is sampled from $\mathcal{U}(0, 0.25)$. On an NVIDIA RTX 4090 GPU, optimization takes 42.52 seconds per image, averaged over 100 samples. For all experiments, we employ a uniform anchor vector across all spatial locations ($\mathbf{a}_{ij} = \mathbf{a}$ for all $i, j$) to simplify the optimization process.

\noindent\textbf{Evaluation protocol.}
We evaluate on the first 100 COCO~\cite{coco} validation images. Tampered regions are generated by applying inpainting within bounding boxes obtained by expanding object segmentation masks by a factor of 1.2. For fair comparison, we adjust the watermark strength of baselines~\cite{wam, omniguard, stableguard} to match our PSNR. For StableGuard~\cite{stableguard}, we calculate the PSNR between the watermarked and original images, rather than VAE reconstructions, to ensure evaluation consistency. We report F1, AUC, and IoU as localization evaluation metrics. \sys refers to training-free prediction, while \syss employs the shallow decoder for fine-grained mask refinement.

\noindent\textbf{Shallow mask decoder.} The optional decoder is trained independently on COCO~\cite{coco} training images using both segmentation and random rectangular masks as supervision. We train for 6 epochs with batch size 64, lr$=3\times10^{-4}$, and image size $256\times256$. We used $\alpha_1=0.3$, $\alpha_2=0.9$, $\beta_1=0.7$, and $\beta_2=1.1$.

\begin{figure*}[t]
  \centering
  \begin{minipage}{0.38\textwidth}
    \centering
    \captionof{table}{Quantitative performance in the fully regenerated (FR) setting. \best{Red}: best, \sbest{Blue}: second-best.}
    \label{tab:performance_fr}
    \resizebox{\linewidth}{!}{%
    \begin{tabular}{@{}lcccc@{}}
      \toprule
       & PSNR & F1 & AUC & IoU \\
      \midrule
      WAM          & 32.30 & 0.91 & \sbest{0.95} & 0.84 \\
      OmniGuard    & \sbest{32.65} & 0.88 & 0.49 & 0.79 \\
      StableGuard  & 25.93 & 0.12 & 0.50 & 0.07 \\
      \rowcolor{ours} \sys         & \best{32.80} & \sbest{0.95} & \best{0.97} & \sbest{0.90} \\
      \rowcolor{ours} \syss & \best{32.80} & \best{0.96} & \best{0.97} & \best{0.92} \\
      \bottomrule
    \end{tabular}%
    }
  \end{minipage}
  \hspace{0.02\textwidth}
  \begin{minipage}{0.54\textwidth}
    \centering
    \captionof{table}{Transferability of localization performance in the fully regenerated (FR) setting. \best{Red}: best, \sbest{Blue}: second-best.}
    \label{tab:transfer_fr}
    \resizebox{\linewidth}{!}{%
    \begin{tabular}{@{}lcccccccc@{}}
      \toprule
       & \multicolumn{2}{c}{BrushNet} & \multicolumn{2}{c}{ControlNet} & \multicolumn{2}{c}{HD-Painter} & \multicolumn{2}{c}{Average} \\
      \cmidrule(lr){2-3} \cmidrule(lr){4-5} \cmidrule(lr){6-7} \cmidrule(l){8-9}
       & AUC & IoU & AUC & IoU & AUC & IoU & AUC & IoU \\
      \midrule
      WAM          & \sbest{0.97} & 0.82 & 0.96 & 0.87 & \best{0.99} & 0.87 & \sbest{0.97} & 0.85 \\
      Omni.        & 0.48 & 0.79 & 0.52 & 0.79 & 0.56 & 0.80 & 0.52 & 0.79 \\
      Stab.        & 0.50 & 0.07 & 0.52 & 0.07 & 0.53 & 0.07 & 0.52 & 0.07 \\
      \rowcolor{ours} \sys         & \sbest{0.97} & \sbest{0.89} & \best{0.99} & \best{0.95} & \sbest{0.97} & \sbest{0.91} & \best{0.98} & \sbest{0.91} \\
      \rowcolor{ours} \syss & \best{0.98} & \best{0.92} & \sbest{0.98} & \sbest{0.94} & \sbest{0.97} & \best{0.92} & \best{0.98} & \best{0.93} \\
      \bottomrule
    \end{tabular}%
    }
  \end{minipage}

  \vspace{0.5em}

  \begin{minipage}{0.98\linewidth}
    \centering
    \includegraphics[width=0.99\linewidth, trim={0.1cm 0.3cm 0.1cm 0cm}, clip]{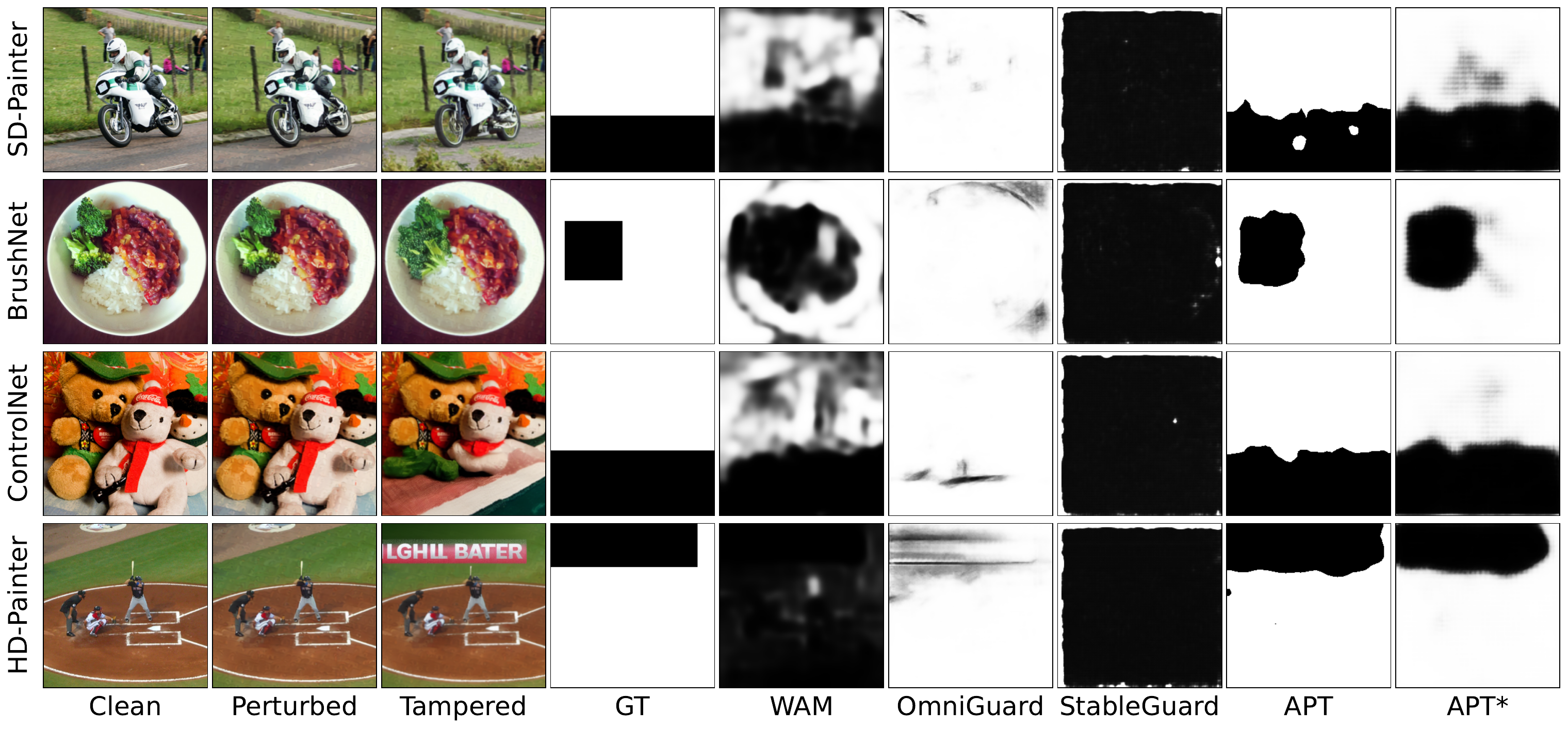}
    \captionof{figure}{Qualitative localization comparison under the FR setting across four inpainting models~\cite{ldm, brushnet, controlnet, hdpainter}. While baselines produce degenerate or near-uniform predictions due to background signal disruption, both \sys and \syss consistently predict accurate masks across diverse inpainting models.}
    \label{fig:qual_fr}
  \end{minipage}

\end{figure*}

\subsection{Localization Performance in the FR Setting}

\noindent\textbf{Evaluation on SD-Painter.}
\cref{tab:performance_fr} reports localization performance on Stable Diffusion Inpainting (SD-Painter~\cite{ldm}) in the FR setting. Existing methods such as StableGuard~\cite{stableguard} and OmniGuard~\cite{omniguard} yield near-random prediction (AUC $\approx 0.5$), indicating that their embedded signals are entirely erased by background regeneration. WAM~\cite{wam} demonstrates modest robustness to FR manipulations, yet its localization remains limited, reaching an IoU of 0.84. In contrast, \syss achieves a superior IoU of 0.92, demonstrating that our vector-wise embedding with latent perturbation preserves localization cues even under the diffusion process.
As shown in \cref{fig:qual_fr}, \sys and \syss consistently yield sharp and spatially accurate predictions, while baselines produce degenerate or uniform masks under FR conditions due to signal loss. 

\noindent\textbf{Transferability to diverse inpainting models.}
We evaluate the generalization of our anchor-aligned perturbation across representative state-of-the-art inpainting models (\eg BrushNet~\cite{brushnet}, ControlNet~\cite{controlnet}, and HD-Painter~\cite{hdpainter}) that utilize diverse background integration strategies detailed in \cref{sec:related_inpainting}. As shown in \cref{tab:transfer_fr}, \syss maintains a consistent average IoU of 0.93, in line with the results in \cref{tab:performance_fr}. This stability stems from leveraging the semi-fragility of aligned vectors rather than visual cues, allowing for consistent localization. While baselines generally follow the failure trends in \cref{tab:performance_fr}, WAM~\cite{wam} shows slightly higher performance on HD-Painter, as HD-Painter tends to generate high-saturation, illustration-like images with distinct boundaries. Overall, these results demonstrate that our anchor-aligned perturbation provides a robust and generalizable localization mechanism across modern diffusion-based inpainting frameworks.

\begin{figure*}[t]
  \centering
  
  \begin{minipage}{0.38\textwidth}
    \centering
    \captionof{table}{Quantitative performance in the spliced (SP) setting. \best{Red}: best, \sbest{Blue}: second-best.}
    \label{tab:performance_sp}
    \resizebox{\linewidth}{!}{%
    \begin{tabular}{@{}lcccc@{}}
      \toprule
       & PSNR & F1 & AUC & IoU \\
      \midrule
      WAM          & 32.30 & \best{1.00} & \best{1.00} & \sbest{0.99} \\
      OmniGuard    & \sbest{32.65} & \best{1.00} & \best{1.00} & \best{1.00} \\
      StableGuard  & 25.93 & \best{1.00} & \best{1.00} & \best{1.00} \\
      \rowcolor{ours} \sys         & \best{32.80} & \sbest{0.97} & \sbest{0.99} & 0.94 \\
      \rowcolor{ours} \syss & \best{32.80} & \sbest{0.97} & \sbest{0.99} & 0.95 \\
      \bottomrule
    \end{tabular}%
    }
  \end{minipage}
  \hspace{0.02\textwidth}
  \begin{minipage}{0.54\textwidth}
    \centering
    \captionof{table}{Transferability across diverse inpainting models under the spliced (SP) setting. \best{Red}: best, \sbest{Blue}: second-best.}
    \label{tab:transfer_sp}
    \resizebox{\linewidth}{!}{%
    \begin{tabular}{@{}lcccccccc@{}}
      \toprule
       & \multicolumn{2}{c}{BrushNet} & \multicolumn{2}{c}{ControlNet} & \multicolumn{2}{c}{HD-Painter} & \multicolumn{2}{c}{Average} \\
      \cmidrule(lr){2-3} \cmidrule(lr){4-5} \cmidrule(lr){6-7} \cmidrule(l){8-9}
       & AUC & IoU & AUC & IoU & AUC & IoU & AUC & IoU \\
      \midrule
      WAM          & \best{1.00} & \sbest{0.99} & \sbest{0.99} & \sbest{0.99} & \best{1.00} & \sbest{0.99} & \best{1.00} & \sbest{0.99} \\
      Omni.        & \best{1.00} & \best{1.00} & \best{1.00} & \best{1.00} & \best{1.00} & \sbest{0.99} & \best{1.00} & \sbest{0.99} \\
      Stab.        & \best{1.00} & \best{1.00} & \best{1.00} & \best{1.00} & \best{1.00} & \best{1.00} & \best{1.00} & \best{1.00} \\
      \rowcolor{ours} \sys         & \sbest{0.99} & 0.95 & \sbest{0.99} & 0.96 & \sbest{0.99} & 0.96 & \sbest{0.99} & 0.96 \\
      \rowcolor{ours} \syss & \sbest{0.99} & 0.95 & \sbest{0.99} & 0.96 & \sbest{0.99} & 0.94 & \sbest{0.99} & 0.95 \\
      \bottomrule
    \end{tabular}%
    }
  \end{minipage}

  \vspace{0.5em} 

  \begin{minipage}{0.98\linewidth}
    \centering
    \includegraphics[width=0.99\linewidth, trim={0.1cm 0.3cm 0.1cm 0cm}, clip]{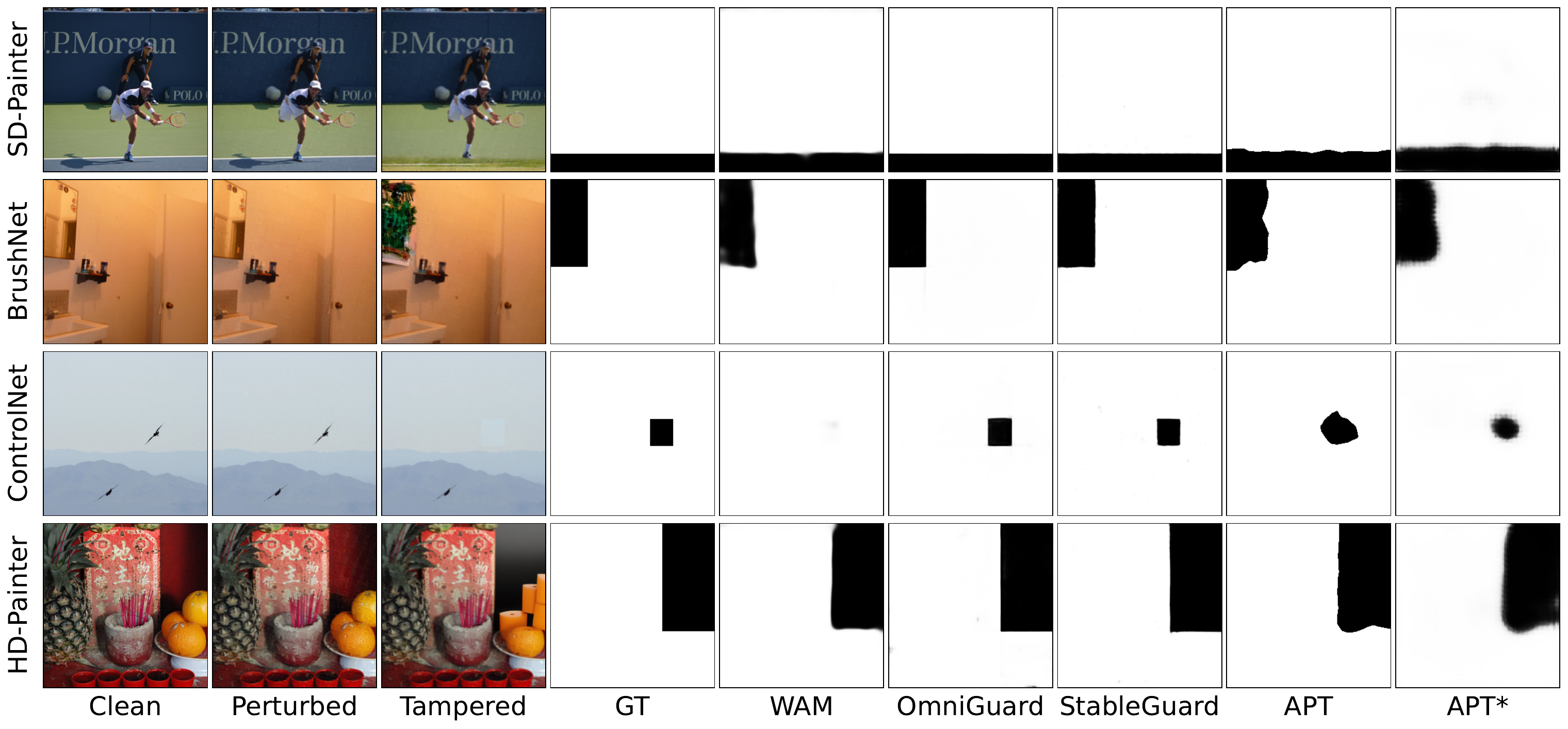}
    \captionof{figure}{Qualitative localization comparison in the SP setting. All methods produce visually plausible masks, and the \sys variants remain highly competitive with state-of-the-art baselines despite not being explicitly optimized for splicing boundaries.}
    \label{fig:qual_sp}
  \end{minipage}

\end{figure*}

\subsection{Localization Performance in the SP Setting}

\noindent\textbf{Evaluation on SD-Painter.}
\cref{tab:performance_sp} presents SP localization results on SD-Painter~\cite{ldm}. In the SP setting, all methods achieve near-perfect performance, and both \sys and \syss maintain a competitive AUC of 0.99, confirming reliable discrimination between perturbed and tampered regions. The marginal IoU gap relative to baselines stems from the fact that our framework is not explicitly optimized for splicing boundaries, but instead captures generalized tampering cues based on anchor alignment. As shown in \cref{fig:qual_sp}, the predicted masks closely follow the ground truth, with discrepancies limited to subtle boundary regions.

\noindent\textbf{Transferability to diverse inpainting models.}
We evaluate the generalization across representative inpainting models~\cite{brushnet, controlnet, hdpainter}. As shown in \cref{tab:transfer_sp}, \sys and \syss consistently achieve an average AUC of 0.99 across all inpainting architectures, confirming that our approach generalizes well to unseen models. Qualitative comparisons across recent inpainting models are provided in \cref{fig:qual_sp}.

\subsection{Robustness to Image Corruptions}
\cref{tab:robust} evaluates robustness under color jitter (10\% brightness increase), Gaussian blur (kernel size 3), and JPEG compression ($Q=95$) in both FR and SP settings. In the FR setting, \syss maintains strong localization with an average IoU of 0.90, demonstrating consistent robustness against diverse image corruptions. In the SP setting, WAM and OmniGuard benefit from seen corruptions during training (All for WAM; JPEG and color jitter for OmniGuard), yet OmniGuard still collapses under JPEG and blur. While both StableGuard and our framework are evaluated on entirely unseen corruptions, StableGuard collapses to an average AUC of 0.80 in the SP setting, whereas both \sys and \syss maintain a robust AUC above 0.96. This demonstrates that the anchor-alignment serves as a robust signal, with the shallow decoder further recovering localization under challenging corruptions.

\begin{table*}[tb]
  \caption{Robustness evaluation under image corruptions. \syss maintains consistent localization against unseen corruptions, with the shallow decoder effectively recovering signal degradation in challenging cases. \best{Red}: best, \sbest{Blue}: second-best.}
  \label{tab:robust}
  \centering
  \resizebox{\textwidth}{!}{
  \begin{tabular}{@{}l cccccccc cccccccc@{}}
    \toprule
     & \multicolumn{8}{c}{Fully Regenerated (FR)} & \multicolumn{8}{c}{Spliced (SP)} \\
    \cmidrule(lr){2-9} \cmidrule(l){10-17}
     & \multicolumn{2}{c}{Color Jitter} & \multicolumn{2}{c}{Gaussian Blur} & \multicolumn{2}{c}{JPEG} & \multicolumn{2}{c}{Average} & \multicolumn{2}{c}{Color Jitter} & \multicolumn{2}{c}{Gaussian Blur} & \multicolumn{2}{c}{JPEG} & \multicolumn{2}{c}{Average} \\
    \cmidrule(lr){2-3} \cmidrule(lr){4-5} \cmidrule(lr){6-7} \cmidrule(lr){8-9} \cmidrule(lr){10-11} \cmidrule(lr){12-13} \cmidrule(lr){14-15} \cmidrule(l){16-17}
     & AUC & IoU & AUC & IoU & AUC & IoU & AUC & IoU & AUC & IoU & AUC & IoU & AUC & IoU & AUC & IoU \\
    \midrule
    WAM         & 0.94 & 0.79 & \sbest{0.95} & 0.86 & \best{0.94} & \sbest{0.80} & \sbest{0.94} & \sbest{0.82} & \best{1.00} & \sbest{0.99} & \best{1.00} & \best{1.00} & \best{0.99} & \best{0.99} & \best{0.99} & \best{0.99} \\
    Omni.       & 0.48 & \sbest{0.79} & 0.51 & 0.79 & 0.47 & 0.79 & 0.49 & 0.79 & \sbest{0.99} & \best{1.00} & 0.58 & 0.79 & 0.47 & 0.79 & 0.82 & \sbest{0.93} \\
    Stab.       & 0.50 & 0.07 & 0.52 & 0.07 & 0.53 & 0.07 & 0.52 & 0.07 & 0.98 & 0.94 & 0.91 & 0.12 & 0.52 & 0.07 & 0.80 & 0.37 \\
    \rowcolor{ours} \sys         & \sbest{0.95} & 0.88 & 0.95 & \sbest{0.87} & 0.90 & 0.64 & 0.93 & 0.80 & 0.98 & 0.91 & 0.97 & 0.92 & 0.94 & 0.78 & 0.96 & 0.87 \\
    \rowcolor{ours} \syss & \best{0.96} & \best{0.91} & \best{0.96} & \best{0.91} & \sbest{0.93} & \best{0.88} & \best{0.95} & \best{0.90} & 0.98 & 0.93 & \sbest{0.97} & \sbest{0.93} & \sbest{0.95} & \sbest{0.91} & \sbest{0.97} & 0.93 \\
    \bottomrule
  \end{tabular}
  }
\end{table*}

\subsection{Alignment Disparity}
\label{exp:fragility}
\begin{figure}[tb]
  \centering
  \begin{minipage}[c]{0.40\linewidth}
    \centering
    \includegraphics[width=0.99\linewidth, trim={0.2cm 0.3cm 0.2cm 0.2cm}, clip]{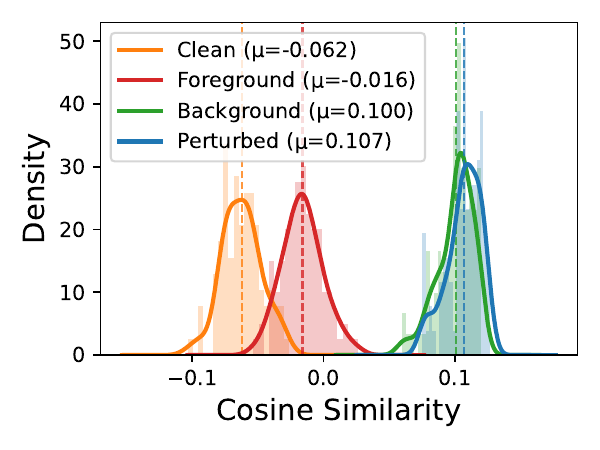}
    \caption{Cosine similarity distributions empirically validating the alignment-based semi-fragility.}
    \label{fig:assumpton}
  \end{minipage}
  \hfill
  \begin{minipage}[c]{0.58\linewidth}
    \centering
    \captionof{table}{Ablation study on each proposed component. $\mathcal{L}_{\text{noise}}$
 and $\mathcal{L}_{\text{hard}}$ provide orthogonal gains across both FR and SP settings, with the full model achieving the best localization performance.}
    \label{tab:ablation}
    \resizebox{\linewidth}{!}{
    \begin{tabular}{@{}cccccccccc@{}}
        \toprule
         & & Perceptual & \multicolumn{3}{c}{FR} & \multicolumn{3}{c}{SP} \\
        \cmidrule(lr){3-3} \cmidrule(lr){4-6} \cmidrule(l){7-9}
        $\mathcal{L}_{\text{noise}}$ & $\mathcal{L}_{\text{hard}}$ & PSNR & F1 & AUC & IoU & F1 & AUC & IoU \\
        \midrule
                                     &                             & \textbf{35.71} & 0.49 & 0.77 & 0.35 & 0.71 & 0.86 & 0.57 \\
        \checkmark                   &                             & 35.22 & 0.72 & 0.86 & 0.59 & 0.88 & 0.93 & 0.80 \\
                                     & \checkmark                  & 34.77 & 0.72 & 0.89 & 0.61 & 0.91 & 0.96 & 0.86 \\
        \checkmark                   & \checkmark                  & 32.80 & \textbf{0.94} & \textbf{0.97} & \textbf{0.90} & \textbf{0.97} & \textbf{0.99} & \textbf{0.94} \\
        \bottomrule
      \end{tabular}
    }
  \end{minipage}
  \vspace{-3mm}
\end{figure}

To empirically validate the alignment disparity discussed in \cref{method:perturb}, we analyze the cosine similarity distributions using 100 COCO validation images. We compare the average similarity scores of perturbed images, clean images, and the foreground and background regions of fully regenerated images tampered with center-mask SD-Painter~\cite{ldm}. As illustrated in \cref{fig:assumpton}, the synthesized foreground regions exhibit a significant deviation from the anchor vectors (mean $\mu$=-0.016), comparable to the low-alignment in clean images ($\mu$= -0.062). In contrast, the background regions of the tampered images maintain a high alignment ($\mu$=0.100) that is nearly identical to that of the perturbed images ($\mu$=0.107). These results demonstrate that foreground regions follow the generative model's native distribution, exhibiting low anchor alignment.

\subsection{Feature Layer Selection}
\label{sec:layer_select}
\begin{figure}[tb]
  \centering
  \begin{minipage}{0.57\linewidth}
    \centering
    \includegraphics[width=0.99\linewidth, trim={0.1cm 0.3cm 0.1cm 0.2cm}, clip]{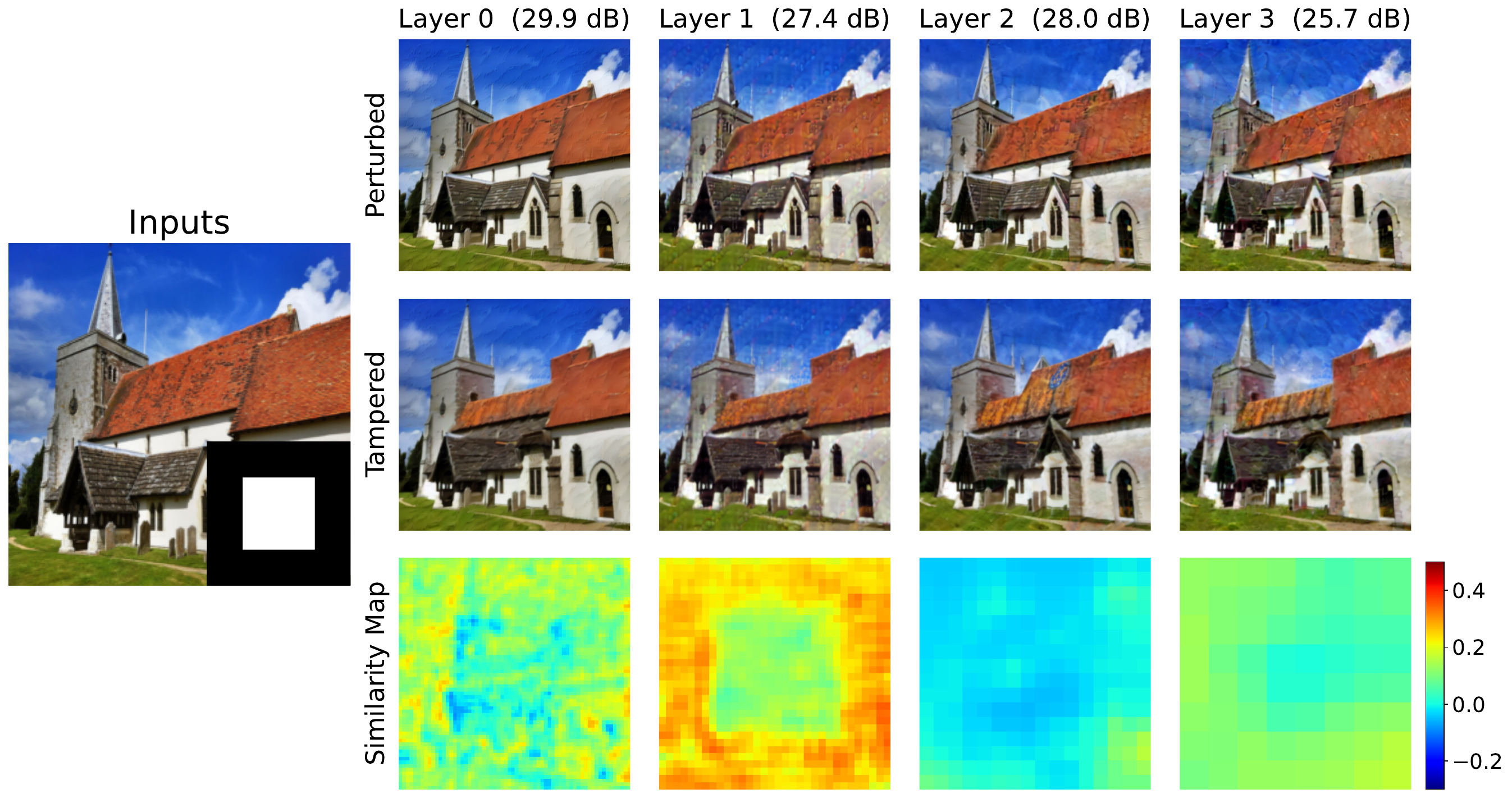}
    \caption{Cosine similarity maps across image encoder feature layers (0-3) under FR inpainting.}
    \label{fig:sim}
  \end{minipage}\hfill
  \begin{minipage}{0.41\linewidth}
    \centering
    \includegraphics[width=0.99\linewidth, trim={0.1cm 0.2cm 0.1cm 0.2cm}, clip]{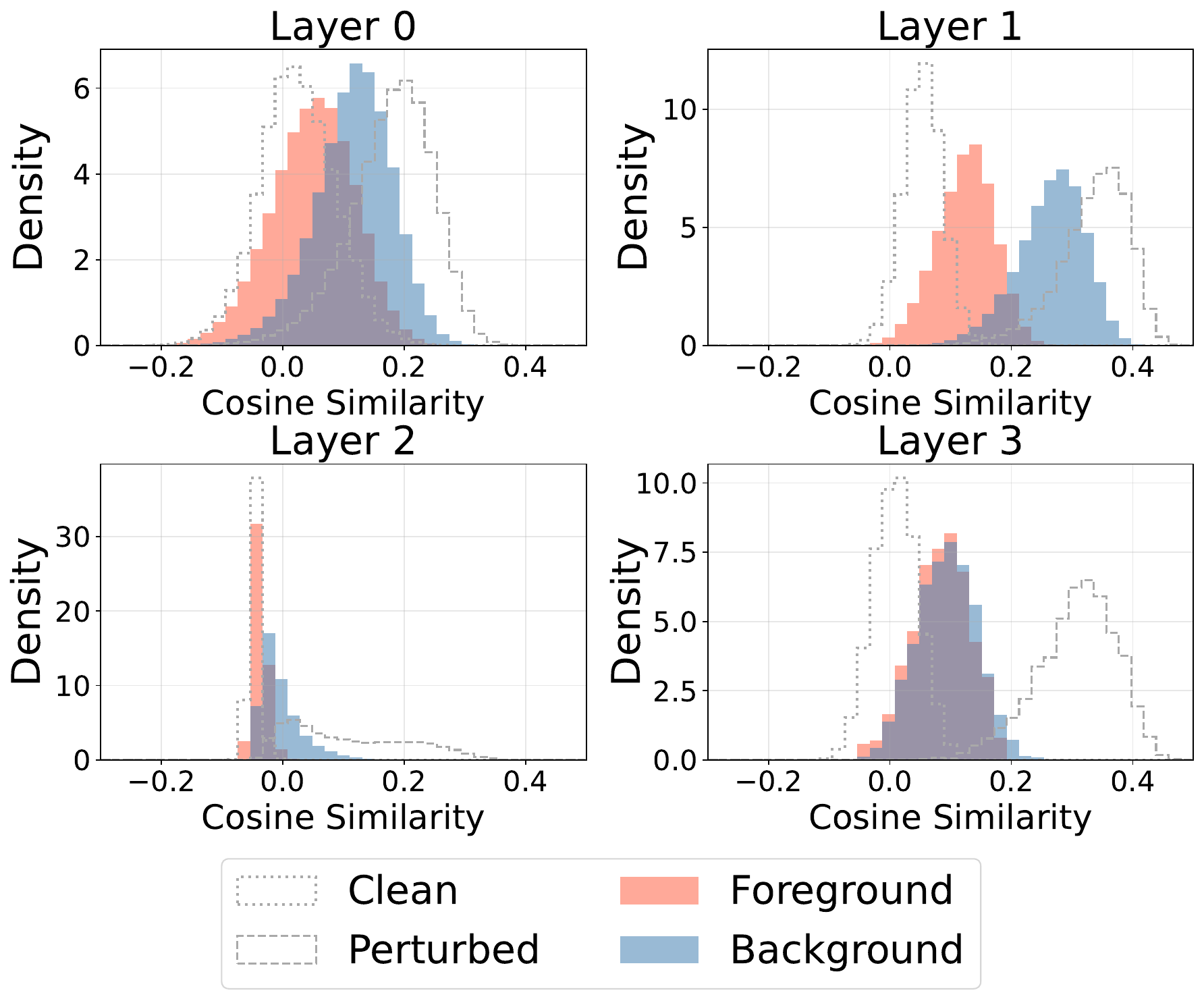}
    \caption{Cosine similarity distributions across feature layers.}
    
    \label{fig:dist}
  \end{minipage}
  \vspace{-3mm}
\end{figure}

To assess the anchor alignment across the image encoder layers, we optimize perturbations on 100 COCO training images and evaluate cosine similarity maps after center-mask SD-Painter FR inpainting. As shown in \cref{fig:sim}, Layer 1 exhibits the sharpest spatial boundary, supported by a distinct bimodal separation between foreground (mean $\mu$=0.124) and background ($\mu$=0.265) in \cref{fig:dist}. We found that Layer 0 produces high-frequency perturbations that are vulnerable to the VAE bottleneck~\cite{ldm} (see Suppl. for details). In deeper layers (2 and 3), wider receptive fields cause foreground tampering cues to entangle with surrounding background vectors, collapsing the foreground-background disparity. Layer 1 is thus selected for its superior balance of robustness and spatial granularity.

\vspace{-3mm}

\subsection{Ablation Study}
\cref{tab:ablation} validates the contribution of each proposed component upon the baseline $\mathcal{L}_{\text{hinge}}(\tilde{\mathbf{x}}_{\text{perturb}}) + \lambda_p \mathcal{L}_{\text{PSNR}} + \lambda_l \mathcal{L}_{\text{LPIPS}}$.
Here, $\mathcal{L}_{\text{noise}} = \mathcal{L}_{\text{hinge}}(\tilde{\mathbf{x}}_{\text{noisy}})$ denotes the alignment on the noisy perturbation branch, and $\mathcal{L}_{\text{hard}}$ denotes the hard negative mining loss on $\tilde{\mathbf{x}}_{\text{perturb}}$. Each component independently improves localization, with $\mathcal{L}_{\text{noise}}$ improving SP and FR IoU by +0.23 and +0.24, and $\mathcal{L}_{\text{hard}}$ providing a larger boost (+0.29 SP, +0.26 FR).
The full model (\cref{eq:full}) achieves the best performance, demonstrating the independent effectiveness of the two components.

\vspace{-3mm}

\section{Conclusion}
We presented \sys, a semi-fragile perturbation framework that addresses the fully regenerated (FR) inpainting setting overlooked by existing proactive localization methods. By embedding dense, vector-wise signals via latent perturbation, \sys achieves alignment-based semi-fragility--robust to background regeneration yet fragile to foreground synthesis. Experiments across a range of inpainting architectures demonstrate that \sys significantly outperforms existing methods in the FR setting while remaining competitive in the SP setting, positioning it as a practical forensic framework generalizable across tampering types unknown at test time.


\section*{Acknowledgements}

This work was supported by Institute of Information \& Communications Technology Planning \& Evaluation (IITP) grant funded by the Korea government (MSIT) (No.RS-2025-25443318, Physically-grounded Intelligence: A Dual Competency Approach to Embodied AGI through Constructing and Reasoning in the Real World), (RS-2023-00237965, Recognition, Action and Interaction Algorithms for Open-world Robot Service), and (No. RS-2020-II200153). Prof. Sung-Eui Yoon and Prof. Sooel Son are co-corresponding authors.

%
%
\bibliographystyle{splncs04}
\bibliography{main}

\clearpage


\setcounter{page}{1}
\setcounter{section}{0}
\renewcommand{\thesection}{\Alph{section}}


\section*{\Large Supplementary Material}

\startcontents[supp]

\noindent{\large\bfseries Table of Contents}\quad
\leaders\hrule height 0.5ex depth -0.46ex\hfill\kern0pt
\par\vspace{0.3em}

\begingroup
\setcounter{tocdepth}{2}
\providecommand\authcount[1]{}
\printcontents[supp]{}{1}{}
\endgroup
\vspace{0.3em}

\noindent\rule{\linewidth}{0.4pt}
\vspace{0.2pt}

In this supplementary material, we provide additional results, analyses, implementation details, and discussions. \cref{supp:results} presents additional evaluations on segmentation masks, 500 inpainting images, and transferability across different VAEs and architectures. \cref{supp:analysis} analyzes baseline signal disruptions and perturbations across encoder layers, and discusses efficiency-quality trade-offs. \cref{supp:details} details the optimization algorithm, mask generation for the shallow decoder, and evaluation protocols. Finally, \cref{supp:discussion} discusses our threat model, limitations, and societal impacts.

Throughout this supplementary, we use \sys to refer to our method with training-free prediction, and \syss to denote the variant that employs the trained shallow mask decoder. The code will be made publicly available.

\setcounter{page}{1}
\setcounter{section}{0}
\renewcommand{\thesection}{\Alph{section}}

\section{Additional Results}
\label{supp:results}

\subsection{Evaluation on Segmentation Masks}

\begin{figure*}[tb]
  \centering
  
  \begin{minipage}{0.38\textwidth}
    \centering
    \captionof{table}{Quantitative performance in the FR setting on segmentation masks. \best{Red}: best, \sbest{Blue}: second-best.}
    \label{tab:performance_segm}
    \resizebox{\linewidth}{!}{%
    \begin{tabular}{@{}lcccc@{}}
      \toprule
       & PSNR & F1 & AUC & IoU \\
      \midrule
      WAM          & 32.30 & 0.93 & \best{0.96} & 0.88 \\
      OmniGuard    & \sbest{32.65} & 0.95 & 0.53 & \sbest{0.91} \\
      StableGuard  & 25.93 & 0.12 & 0.46 & 0.06 \\
      \rowcolor{ours} \sys         & \best{32.80} & \sbest{0.96} & \best{0.96} & 0.92 \\
      \rowcolor{ours} \syss & \best{32.80} & \best{0.97} & \sbest{0.95} & \best{0.94} \\
      \bottomrule
    \end{tabular}%
    }
  \end{minipage}
  \hspace{0.02\textwidth}
  \begin{minipage}{0.54\textwidth}
    \centering
    \captionof{table}{Transferability of localization performance in the FR setting on segmentation masks. \best{Red}: best, \sbest{Blue}: second-best.}
    \label{tab:transfer_segm}
    \resizebox{\linewidth}{!}{%
    \begin{tabular}{@{}lcccccccc@{}}
      \toprule
       & \multicolumn{2}{c}{BrushNet} & \multicolumn{2}{c}{ControlNet} & \multicolumn{2}{c}{HD-Painter} & \multicolumn{2}{c}{Average} \\
      \cmidrule(lr){2-3} \cmidrule(lr){4-5} \cmidrule(lr){6-7} \cmidrule(l){8-9}
       & AUC & IoU & AUC & IoU & AUC & IoU & AUC & IoU \\
      \midrule
      WAM          & \sbest{0.96} & 0.87 & 0.95 & 0.89 & \best{0.98} & 0.88 & \sbest{0.96} & 0.88 \\
      Omni.        & 0.58 & \sbest{0.91} & 0.52 & 0.90 & 0.55 & 0.90 & 0.55 & 0.90 \\
      Stab.        & 0.47 & 0.06 & 0.50 & 0.06 & 0.51 & 0.06 & 0.49 & 0.06 \\
      \rowcolor{ours} \sys         & \best{0.97} & \sbest{0.91} & \best{0.98} & \best{0.96} & \sbest{0.95} & \sbest{0.92} & \sbest{0.96} & \sbest{0.93} \\
      \rowcolor{ours} \syss & \best{0.97} & \best{0.94} & \best{0.98} & \best{0.96} & \sbest{0.95} & \best{0.94} & \best{0.97} & \best{0.95} \\
      \bottomrule
    \end{tabular}%
    }
  \end{minipage}

  \vspace{0.5em}

  \begin{minipage}{0.98\linewidth}
    \centering
    \includegraphics[width=0.99\linewidth, trim={0.1cm 0.3cm 0.1cm 0cm}, clip]{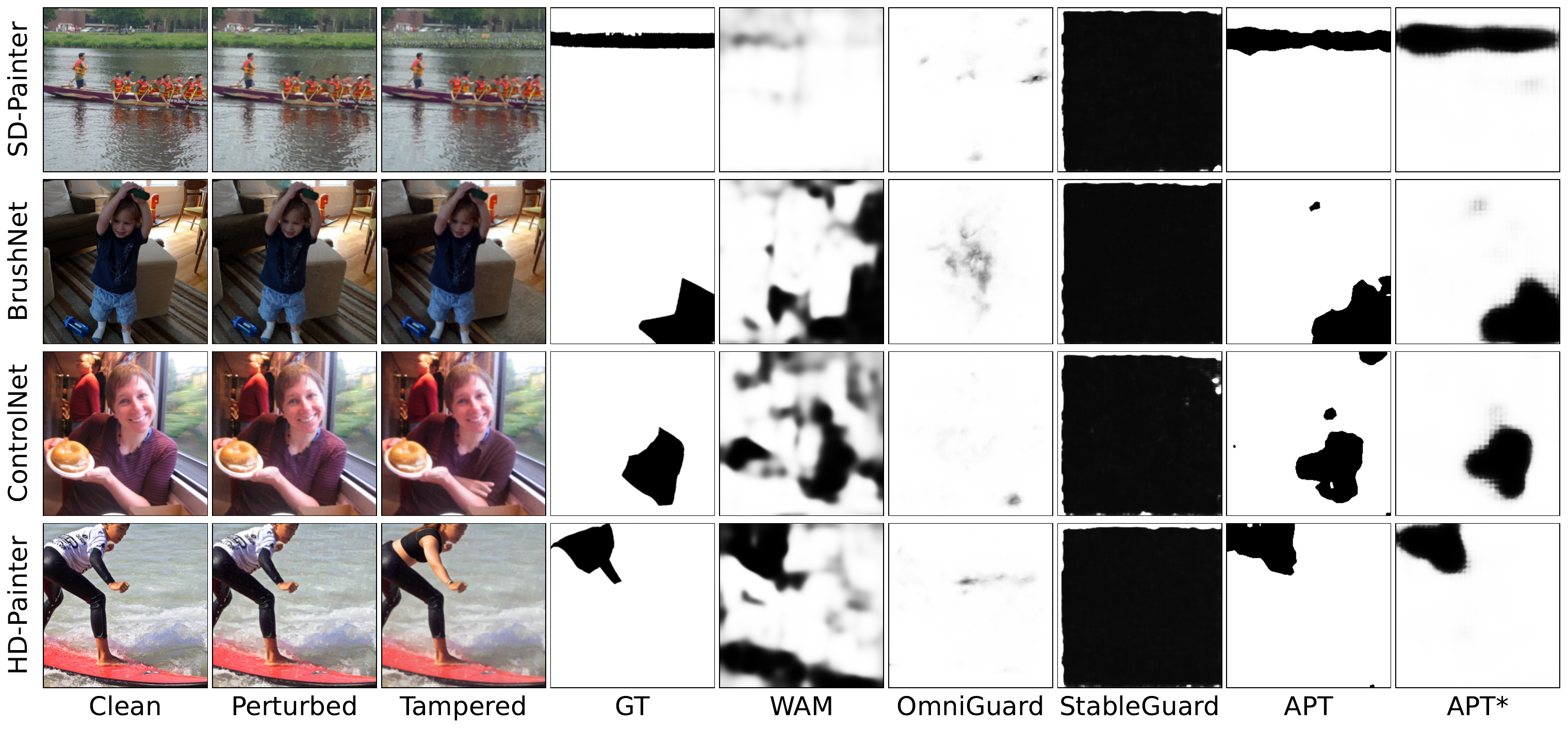}
    \captionof{figure}{Qualitative localization comparison under the FR setting on segmentation masks across four inpainting models.}
    \label{fig:qual_segm}
  \end{minipage}

\end{figure*}

We further evaluate localization performance on images manipulated using instance segmentation masks. \cref{tab:performance_segm} (for SD-Painter) and \cref{tab:transfer_segm} (across four inpainting models) show that both \sys and \syss consistently outperform all baselines in the challenging fully regenerated (FR) setting. Qualitative results in \cref{fig:qual_segm} further demonstrate the superior localization results of our framework.

\subsection{Evaluation on 500 Images}
We scale our evaluation to 500 inpainting samples across both FR and SP settings, utilizing the first 500 COCO~\cite{coco} validation dataset following prior work~\cite{editguard, omniguard}. \cref{tab:performance_500} shows that \sys and \syss maintain consistent localization, confirming that our localization performance generalizes to larger test sets.

\subsection{Transferability across Different VAEs and Architectures}
We further evaluate the transferability of \syss when the protection perturbation is optimized with a specific SD-VAE but applied to inpainting models with different VAEs or generation architectures.
To this end, we test \syss on five additional inpainters covering diverse model families: a LoRA-fine-tuned SD model (LCM-LoRA-SDv1-5~\cite{lcm-lora}), SDXL-inpainting~\cite{sdxl}, a flow-matching inpainting model (SDv3~\cite{sdv3}), and recent flow-matching T2I models used for inpainting through LanPaint~\cite{lanpaint}, including FLUX.2-klein-9B~\cite{flux2} and Z-Image-Turbo~\cite{zimage}.
Although the perturbation $\boldsymbol{\delta}$ is optimized for a particular SD-VAE, \cref{tab:more_transfer} shows that \syss transfers effectively across non-identical VAEs and recent inpainting architectures, achieving AUC $\geq 0.97$ and IoU $\geq 0.91$.
These results indicate that \syss is not limited to the protection-time SD-VAE and can generalize to non-identical VAEs and deterministic ODE-based generation pipelines.

\begin{table*}[tb]
  \caption{Localization performance on 500 images under both FR and SP settings. \best{Red}: best, \sbest{Blue}: second-best.}
  \label{tab:performance_500}
  \centering
  \resizebox{0.8\textwidth}{!}{
  \begin{tabular}{@{}l c ccc ccc@{}}
    \toprule
     & Perceptual & \multicolumn{3}{c}{Fully Regenerated (FR)} & \multicolumn{3}{c}{Spliced (SP)} \\
    \cmidrule(lr){2-2} \cmidrule(lr){3-5} \cmidrule(l){6-8}
     & PSNR ($\uparrow$) & F1 ($\uparrow$) & AUC ($\uparrow$) & IoU ($\uparrow$) & F1 ($\uparrow$) & AUC ($\uparrow$) & IoU ($\uparrow$) \\
    \midrule
    WAM         & 32.64 & 0.89 & 0.95 & 0.83 & \best{1.00} & \best{1.00} & \sbest{0.99} \\
    OmniGuard       & \sbest{33.17} & 0.86 & 0.50 & 0.77 & \sbest{0.99} & \best{1.00} & \sbest{0.99} \\
    StableGuard       & 26.61 & 0.12 & 0.50 & 0.07 & \best{1.00} & \best{1.00} & \best{1.00} \\
    \rowcolor{ours} \sys         & \best{33.44} & \sbest{0.93} & \sbest{0.97} & \sbest{0.89} & 0.97 & \sbest{0.99} & 0.94 \\
    \rowcolor{ours} \syss & \best{33.44} & \best{0.95} & \best{0.98} & \best{0.91} & 0.97 & \sbest{0.99} & 0.95 \\
    \bottomrule
  \end{tabular}
  }
\end{table*}

\begin{table*}[tb]
  \caption{Transferability performance in the fully regenerated (FR) setting. Results are reported as AUC/IoU. We use the same SD-VAE-optimized perturbation $\boldsymbol{\delta}$ and the same shallow decoder for \syss.}
  \label{tab:more_transfer}
  \centering
  \setlength{\tabcolsep}{6pt}
  \resizebox{0.85\textwidth}{!}{
  \begin{tabular}{l c c c}
    \toprule
    Inpainting Method & VAE Type $(H{=}W, C)$ & APT ($\uparrow$) & APT$^\ast$ ($\uparrow$) \\
    \midrule
    \multicolumn{4}{@{}l}{\textit{Fine-tuned SD}} \\
    \quad LCM-LoRA ('23) & SD-VAE (64, 4) & 0.99/0.94 & 0.98/0.94 \\
    \midrule
    \multicolumn{4}{@{}l}{\textit{Non-identical VAEs}} \\
    \quad SDXL-inpaint ('23) & SDXL-VAE (128, 4) & 0.99/0.94 & 0.98/0.95 \\
    \midrule
    \multicolumn{4}{@{}l}{\textit{Flow-matching models}} \\
    \quad SDv3-inpaint ('24) & SD3-AE (128, 16) & 0.97/0.88 & 0.97/0.91 \\
    \quad FLUX.2 ('25) & FLUX.2-AE (128, 32) & 0.97/0.92 & 0.97/0.94 \\
    \quad Z-Image-Turbo ('25) & FLUX-AE (128, 16) & 0.99/0.95 & 0.98/0.96 \\
    \bottomrule
  \end{tabular}
  }
\end{table*}

\section{Additional Analysis}
\label{supp:analysis}

\subsection{Signal Disruption across Baselines}

\begin{figure}[tb]
  \centering
  \begin{minipage}{0.55\linewidth}
    \centering
    \includegraphics[width=\linewidth]{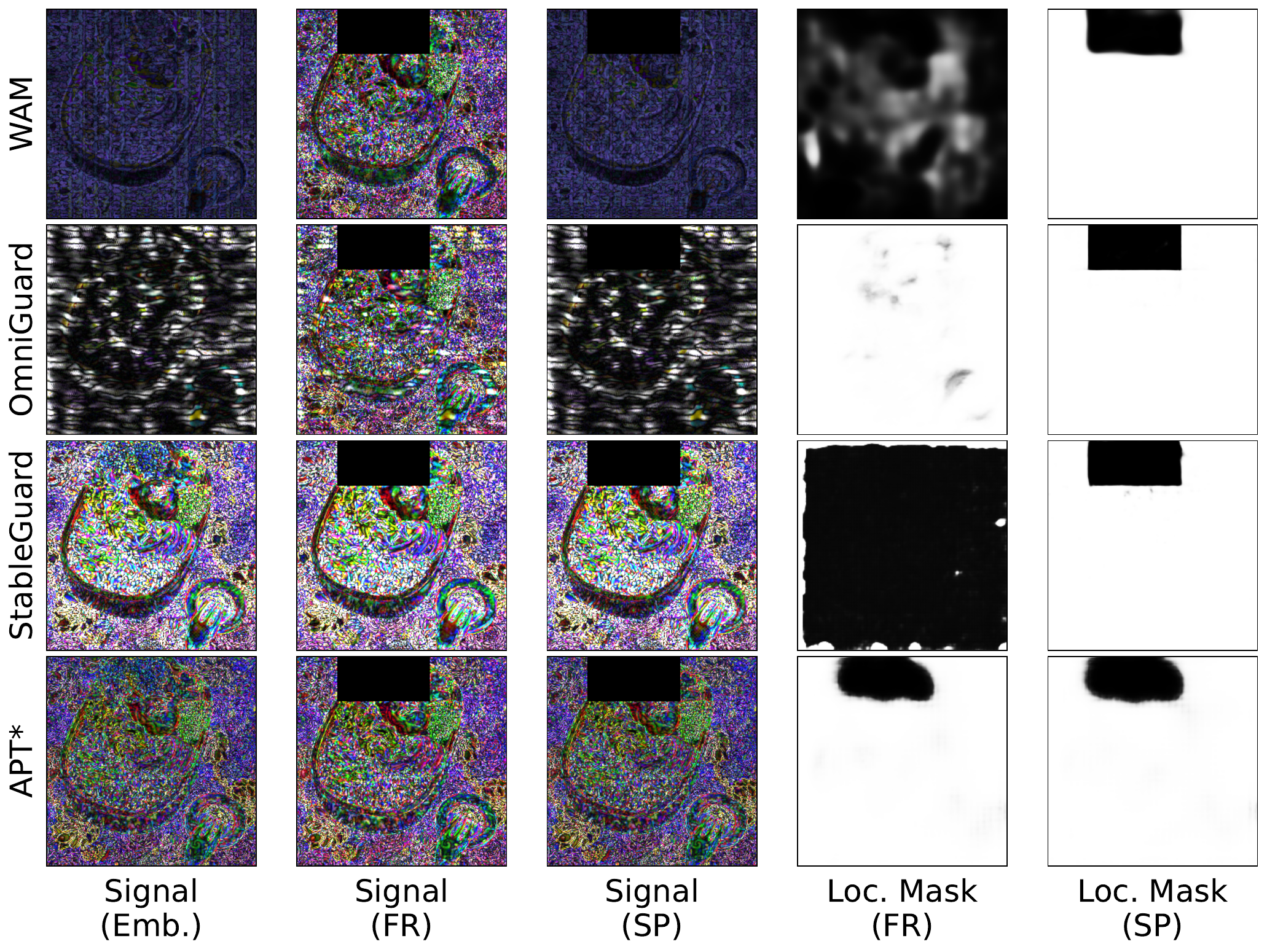}
    \caption{Extended comparison of embedded signals and localization masks across all baselines under FR and SP settings (cf. \cref{fig:fig1}).}
    \label{fig:fig1_all}
  \end{minipage}
  \hfill
  \begin{minipage}{0.43\linewidth}
    \centering
    \includegraphics[width=\linewidth]{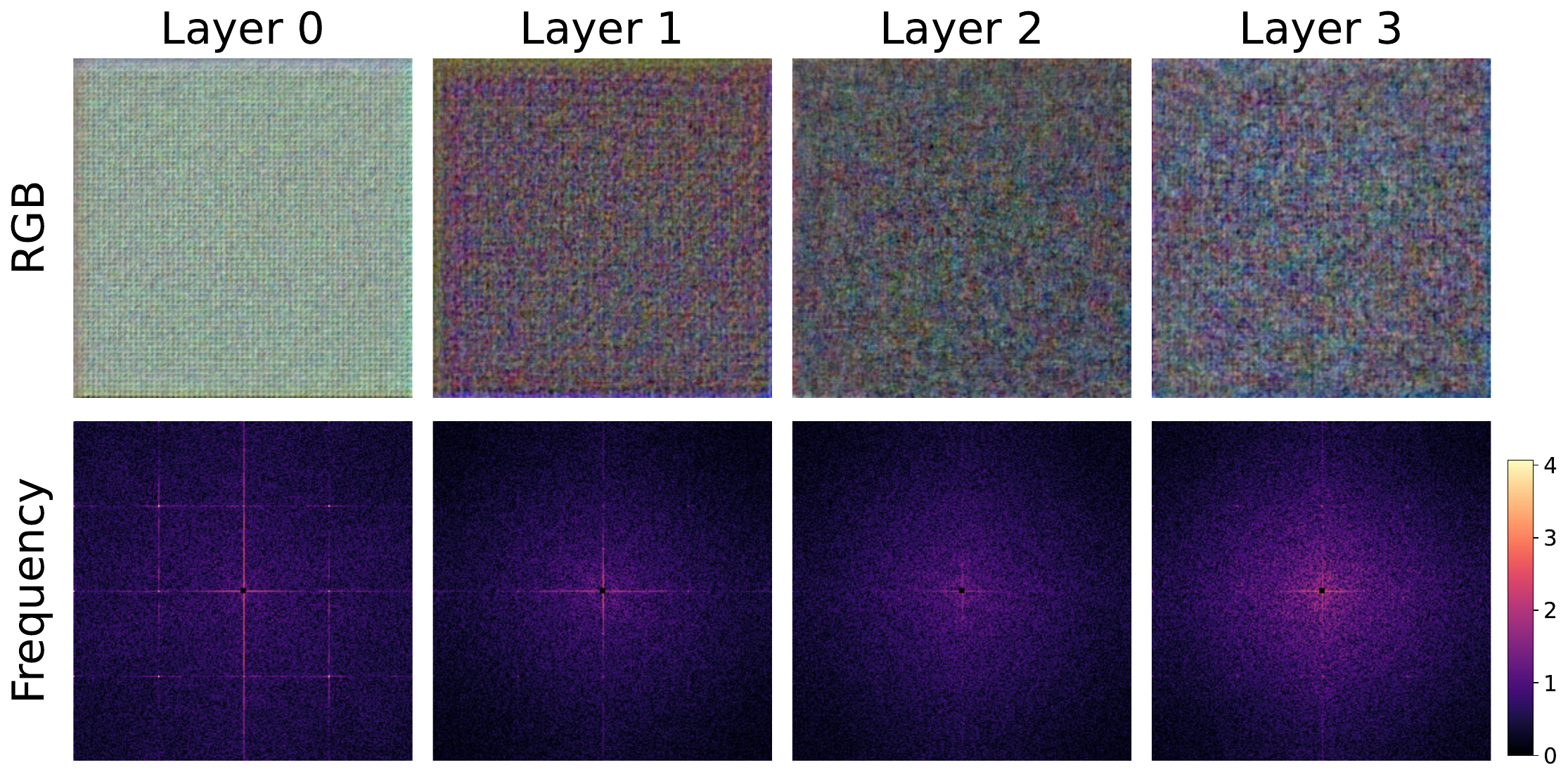}
    \caption{Average perturbations visualized in RGB and frequency domains across encoder layers 0-3.}
    \label{fig:layer_perturb}
  \end{minipage}
\end{figure}

While \cref{fig:fig1} in the main paper visualizes the embedding and background signals for WAM~\cite{wam} and \syss, we extend this comparison to other baselines here. \cref{fig:fig1_all} shows that methods like WAM~\cite{wam} and OmniGuard~\cite{omniguard}, which embed their signals in the pixel and frequency domains, undergo severe alterations in their localization cues in the FR setting. Although both StableGuard~\cite{stableguard} and \syss maintain their signals across both SP and FR settings by leveraging latent space embedding, StableGuard struggles with precise localization due to its reliance on a global embedding, whereas our dense vector-wise approach ensures accurate spatial boundaries.

\subsection{Layer-wise Perturbation Analysis}

Expanding on the discussion in \cref{sec:layer_select}, we further analyze the perturbations across different image encoder layers. Specifically, we optimize perturbations using only the alignment loss (without the hinge) and perceptual losses  to maximize the perturbation strength. The average perturbations in both the RGB and frequency domains are visualized in \cref{fig:layer_perturb}. Layer 0 exhibits fine-grained horizontal and vertical patterns, which manifest as distinct crosses in the high-frequency spectrum. Layer 1 also forms a cross, but its patterns are coarser and predominantly low-frequency. Layers 2 and 3 concentrate their energy almost exclusively in the low-frequency spectrum, lacking distinct edge patterns in the RGB domain. Layer 0 perturbations are dominated by high frequencies that struggle to propagate the VAE bottleneck~\cite{ldm}, while the wider receptive fields of Layers 2–3 hinder precise boundary localization. Layer 1 strikes the right balance, producing coarse structural patterns that survive the VAE and retain sufficient spatial resolution.

\subsection{Trade-offs}

\begin{figure}[tb]
  \centering
  \includegraphics[width=0.98\linewidth, trim={0cm 0cm 0cm 0cm}, clip]{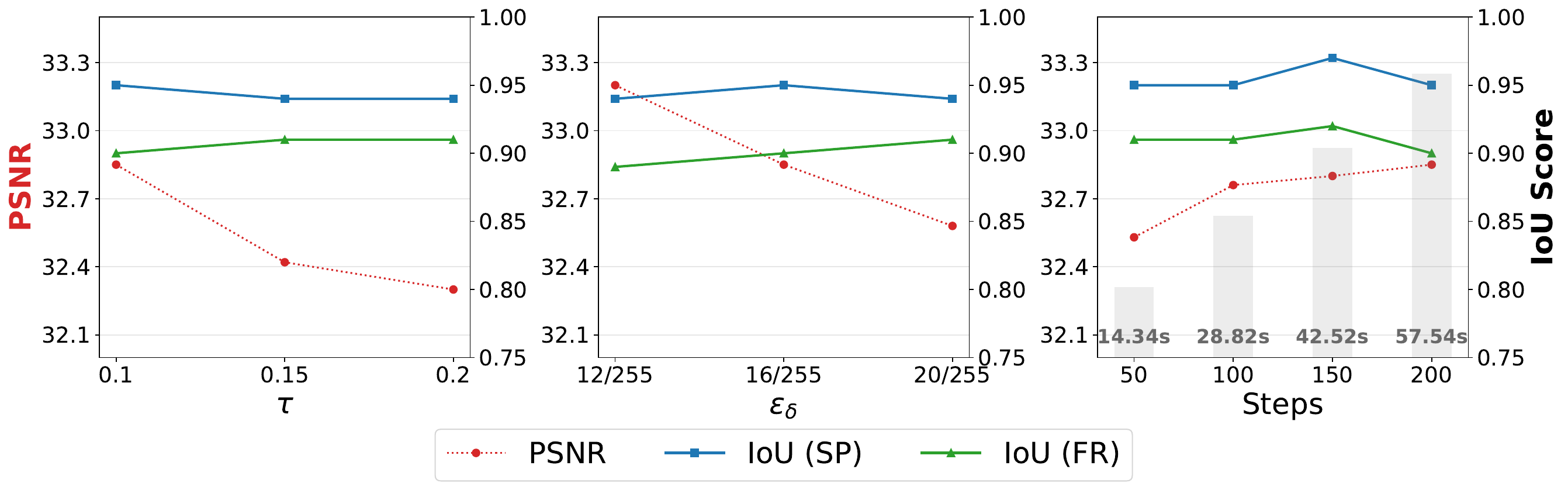}
  \caption{Trade-off analysis varying target cosine similarity $\tau$ (left), perturbation budget $\epsilon_\delta$ (middle), and  optimization steps (right).}
  \label{fig:tradeoff}
\end{figure}

We investigate the trade-offs among computational efficiency, visual quality (PSNR), and localization performance (see \cref{fig:tradeoff}). First, varying the target cosine similarity $\tau$ shifts the PSNR-IoU balance predictably (left). Second, tightening the perturbation budget $\epsilon_\delta$ improves perceptual quality at a modest cost to localization accuracy (middle). Finally, reducing the optimization steps to 50 accelerates the process to 14.34 seconds per image while maintaining highly competitive IoU scores in both settings (right). These results suggest that \syss flexibly adapts to diverse computational and perceptual constraints without substantial performance loss.

\section{Implementation Details}
\label{supp:details}

\subsection{Optimization}
We detail the complete procedure for our anchor-aligned perturbation optimization in \cref{alg:embedding}.

\subsection{Shallow Decoder}

\noindent\textbf{Mask Generation.}
During the training of the shallow mask decoder on the COCO dataset, we employ two mask generation strategies with equal probability: (i) unpaired masks and (ii) random rectangular masks. For unpaired masks, we randomly sample an instance segmentation mask from a different image in the dataset. For random rectangular masks, we generate a bounding box with its width and height uniformly sampled from 10\% to 50\% of the image dimensions. Additionally, we invert the generated mask with a 50\% probability. Then, we define the ground-truth mask as $m_\text{GT} = 1 - \text{mask}$ to represent perturbed regions as 1 and tampered regions as 0.

\begin{algorithm}[tb]
\caption{APT Localization-Mark Embedding Optimization}
\label{alg:embedding}
\textbf{Input:} Original image $x$, Pretrained VAE Encoder $\mathcal{E}_{vae}$, VAE Decoder $\mathcal{D}_{vae}$, Image Encoder $\mathcal{E}_{img}$, Perturbation budget $\epsilon_{\delta}$, Optimization steps $Steps$ \\
\textbf{Output:} Protected image $\tilde{x}_\text{perturb}$
\begin{algorithmic}[1]
\STATE $z \leftarrow \mathcal{E}_{vae}(x)$
\STATE Initialize learnable latent perturbation $\delta \leftarrow \mathbf{0}$
\FOR{$t = 1$ to $Steps$}
    \STATE Sample random binary rectangular mask $m_{rand}$
    \STATE Sample noise scale $\sigma \sim \mathcal{U}(0, 0.25)$
    \STATE Sample Gaussian noise $\nu \sim \mathcal{N}(0, \sigma^2)$
    
    \STATE $\tilde{z} \leftarrow z + \delta$
    \STATE $\tilde{z}_{noisy} \leftarrow \tilde{z} \odot (1 - m_{rand}) + (\tilde{z} + \nu) \odot m_{rand}$
    
    \STATE $\tilde{x}_{perturb} \leftarrow \mathcal{D}_{vae}(\tilde{z})$
    \STATE $\tilde{x}_{noisy} \leftarrow \mathcal{D}_{vae}(\tilde{z}_{noisy})$
    
    \STATE Compute $\mathcal{L}_{hinge}(\hat{x})$ and $\mathcal{L}_{hard}(\hat{x})$ for $\hat{x} \in \{\tilde{x}_{perturb}, \tilde{x}_{noisy}\}$
    \STATE Compute perceptual losses $\mathcal{L}_{PSNR}(\tilde{x}_{perturb}, x)$ and $\mathcal{L}_{LPIPS}(\tilde{x}_{perturb}, x)$
    
    \STATE $\mathcal{L}_{total} \leftarrow \sum_{\hat{x} \in \{\tilde{x}_{perturb}, \tilde{x}_{noisy}\}} \left[ \mathcal{L}_{hinge}(\hat{x}) + \mathcal{L}_{hard}(\hat{x}) \right] + \lambda_p \mathcal{L}_{PSNR} + \lambda_l \mathcal{L}_{LPIPS}$
    \STATE Update $\delta$ by minimizing $\mathcal{L}_{total}$
\ENDFOR
\STATE $\tilde{x} \leftarrow \mathcal{D}_{vae}(z + \delta)$
\STATE $\tilde{x}_{perturb} \leftarrow \text{Clip}(x + \text{Clip}(\tilde{x}- x, -\epsilon_{\delta}, \epsilon_{\delta}), 0, 1)$
\RETURN $\tilde{x}_{perturb}$
\end{algorithmic}
\end{algorithm}

\subsection{Evaluation}

\noindent\textbf{Metrics.} We evaluate pixel-level localization performance using the F1 score, Area Under the Curve (AUC), and Intersection over Union (IoU). F1 and IoU measure the absolute overlap between the predicted and ground-truth masks at a fixed threshold (0.5). In contrast, AUC is a threshold-independent metric that evaluates the model's inherent ability to separate tampered and perturbed pixels. 

This distinction is crucial due to the class imbalance in manipulated images, where the tampered region is often much smaller than the intact background. For instance, a collapsed model that blindly predicts the entire image as the majority class (perturbed) will yield a near-random AUC ($\approx 0.5$) but may produce misleadingly high F1 and IoU scores. Therefore, AUC is essential to verify the model's true discriminative power.

While AUC indicates discriminative power, a high IoU ensures practical usability. In real-world forensic analysis, verifiers rarely know in advance whether an image has been spliced or fully regenerated. A model might achieve high AUC in both settings by separating the classes well internally, but if its output score distributions shift drastically between SP and FR, a single fixed threshold will yield suboptimal results. Thus, a robust forensic framework must deliver high IoU at a constant, predefined threshold across both tampering types. \sys and \syss maintain consistent output probability distributions across both settings (see \cref{tab:performance_fr} and \cref{tab:performance_sp}), allowing a single threshold to generalize effectively.

\noindent\textbf{Other baselines.}
To ensure a fair comparison, we adjust the embedding strengths of the baseline methods to align their perceptual quality with ours. Specifically, we set the watermark strength to 3.0 and 2.0 for WAM~\cite{wam} and OmniGuard~\cite{omniguard}, respectively. Furthermore, StableGuard~\cite{stableguard} originally reports PSNR between the VAE-reconstructed image and the watermarked image, whereas we strictly compute the PSNR directly between the original clean image and the protected image across all methods. This ensures a consistent and practical evaluation of perceptual distortion from embedding signals.

\noindent\textbf{Inpainting models.}
All evaluated models~\cite{ldm, controlnet, hdpainter, brushnet} synthesize images using 50 inference steps and an empty text prompt to fill the masked regions to blend naturally with the surrounding context.

\section{Discussion}
\label{supp:discussion}

\subsection{Threat Model}
We consider three distinct entities: the \textit{image owner} (defender), the \textit{attacker}, and the \textit{verifier}. The image owner protects the original image by injecting an imperceptible perturbation prior to public distribution. To model a practical defense scenario, we assume the detector operates as a black-box API. While the detection algorithm is publicly accessible for integrity verification, the internal anchor vectors remain securely hidden on the server side as a secret key. This separation prevents attackers from trivially extracting the anchors or bypassing the detection mechanism.

\subsection{Limitation}

\sys relies on the alignment disparity between the protected background and the synthesized foreground. If an adversary performs a copy-move forgery within the same protected image, the copied region retains the embedded anchor alignment. This lack of alignment disparity can potentially lead to false negatives. 

\subsection{Societal Impact}

\sys expands the scope of proactive forensics by addressing the fully regenerated setting, a challenging yet realistic scenario in state-of-the-art localized image synthesis. By providing a reliable baseline for this setting, we contribute to the broader effort to ensure practical and robust image authentication in the wild.

However, a potential negative impact exists if the framework is misused. A malicious actor could apply the \sys perturbation to an already-manipulated image, embedding a valid localization signal that misleads the verification system into authenticating the forgery as an intact original. To mitigate this risk, future research should explore integrating proactive defenses directly at the point of capture (\eg, camera hardware).

\end{document}